\pdfoutput=1

\documentclass[10pt,twocolumn,letterpaper]{article}

\usepackage[pagenumbers]{cvpr} % To force page numbers, e.g. for an arXiv version

\usepackage{graphicx}
\usepackage{amsmath}
\usepackage{amssymb}
\usepackage{tabularx}
\usepackage{booktabs}
\usepackage{bbold}

\usepackage{pifont}

\renewenvironment{quote}
  {\list{}{\rightmargin=0.4cm \leftmargin=0.4cm}%
  \item\relax}
  {\endlist}
  
\usepackage[pagebackref,breaklinks,colorlinks]{hyperref}

\usepackage[accsupp]{axessibility}  % Improves PDF readability for those with disabilities.

\usepackage[capitalize]{cleveref}
\crefname{section}{Sec.}{Secs.}
\Crefname{section}{Section}{Sections}
\Crefname{table}{Table}{Tables}
\crefname{table}{Tab.}{Tabs.}

\def\cvprPaperID{6270} % *** Enter the CVPR Paper ID here
\def\confName{CVPR}
\def\confYear{2022}

\begin{document}

\title{Quantifying Societal Bias Amplification in Image Captioning}

% \author{Yusuke Hirota \quad Yuta Nakashima \quad  Noa Garcia\\
% Institute for Datability Science, Osaka University\\
% %Institution1 address\\
% {\tt\small y-hirota@is.ids.osaka-u.ac.jp, \{n-yuta,noagarcia\}@ids.osaka-u.ac.jp}
% %\and
% %Second Author\\
% %Institution2\\
% %First line of institution2 address\\
% %{\tt\small secondauthor@i2.org}
% }

% \author{Yusuke Hirota
% \qquad Yuta Nakashima
% \qquad Noa Garcia  \\
% Osaka University  \\
% {\tt\small y-hirota@is.ids.osaka-u.ac.jp}
% \quad{\tt\small n-yuta@ids.osaka-u.ac.jp}
% \quad{\tt\small noagarcia@ids.osaka-u.ac.jp} \\
% }

\author{Yusuke Hirota\\
Osaka University  \\
{\tt\small y-hirota@is.ids.osaka-u.ac.jp}
\and
Yuta Nakashima \\
Osaka University  \\
{\tt\small n-yuta@ids.osaka-u.ac.jp}
\and
Noa Garcia  \\
Osaka University  \\
{\tt\small noagarcia@ids.osaka-u.ac.jp} \\
}
\maketitle

%%%%%%%%% ABSTRACT
\begin{abstract}
We study societal bias amplification in image captioning. Image captioning models have been shown to perpetuate gender and racial biases, however, metrics to measure, quantify, and evaluate the societal bias in captions are not yet standardized. We provide a comprehensive study on the strengths and limitations of each metric, and propose LIC, a metric to study captioning bias amplification. We argue that, for image captioning, it is not enough to focus on the correct prediction of the protected attribute, and the whole context should be taken into account. We conduct extensive evaluation on traditional and state-of-the-art image captioning models, and surprisingly find that, by only focusing on the protected attribute prediction, bias mitigation models are unexpectedly amplifying bias.
\end{abstract}

%%%%%%%%% BODY TEXT
\section{Introduction}
\label{sec:intro}

The presence of undesirable biases in computer vision applications is of increasing concern. The evidence shows that large-scale datasets, and the models trained on them, present major imbalances in how different subgroups of the population are represented \cite{excavating,buolamwini2018gender,zhao2017men,burns2018women}. Detecting and addressing these biases, often known as societal biases, has become an active research direction in our community \cite{wang2020revise,yang2020towards,khan2021one,de2019does,stock2018convnets,alvi2018turning,thong2021feature}.

Contrary to popular belief, the presence of bias in datasets is not the only cause of unfairness \cite{d2020data}. Model choices and how the systems are trained also have a large impact on the perpetuation of societal bias. This is supported by evidence: 1) models are not only reproducing the inequalities of the datasets but amplifying them \cite{zhao2017men}, and 2) even when trained on balanced datasets, models may still be biased \cite{wang2019balanced} as the depth of historical discrimination is more profound than what it can be manually annotated, \ie, bias is not always evident to the human annotator eye. 

\begin{figure}[t]
  \centering
  \vspace{-5pt}
  \includegraphics[clip, width=0.98\columnwidth]{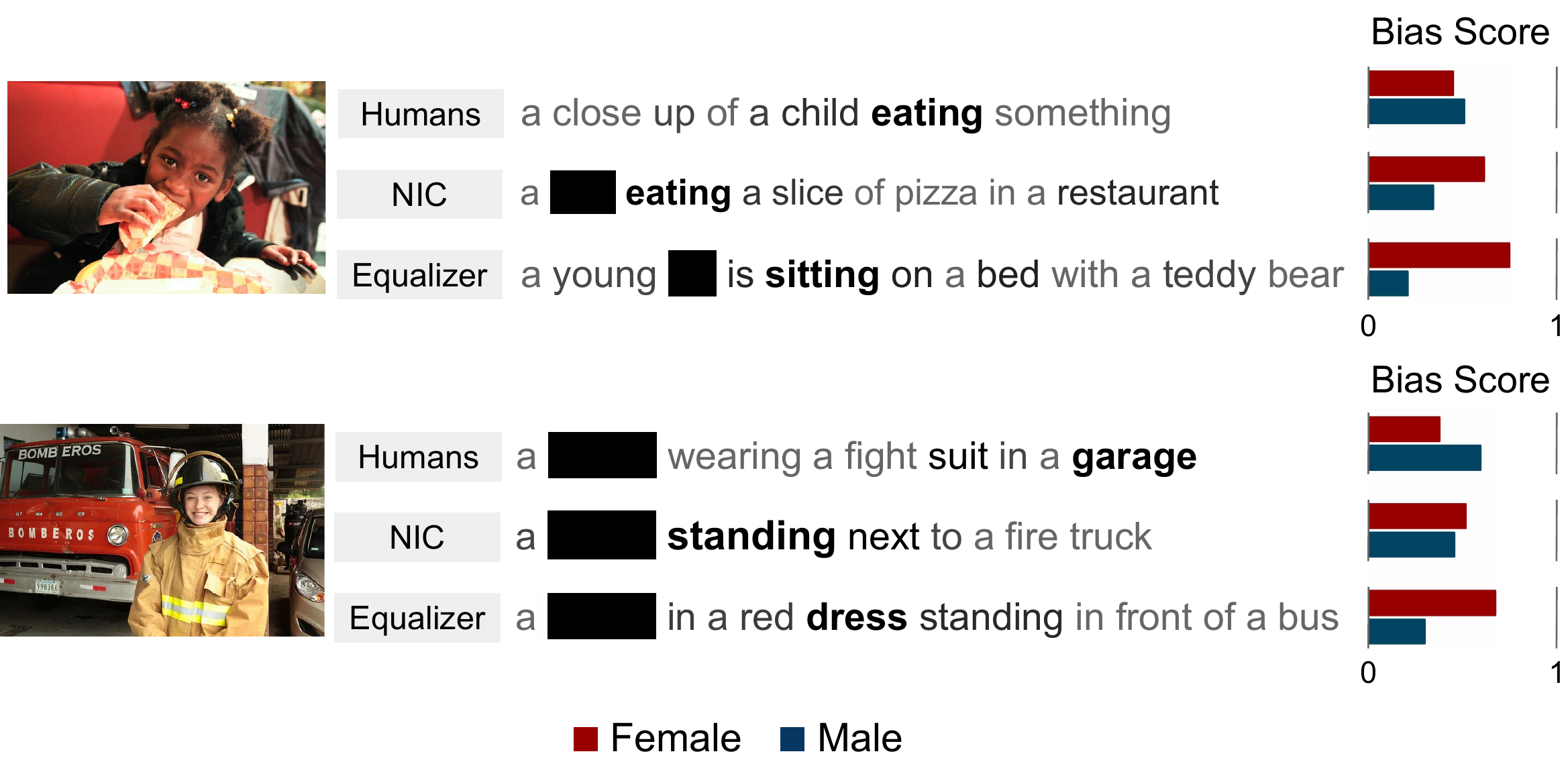}
    \vspace{-5pt}
  \caption{Measuring gender bias in MSCOCO captions \cite{chen2015microsoft}. For each caption generated by humans, NIC \cite{vinyals2015show}, or NIC+Equalizer \cite{burns2018women}, we show our proposed bias score for \textit{female} and \textit{male} attributes. This bias score indicates how much a caption is biased toward a certain protected attribute. The contribution of each word to the bias score is shown in gray-scale (bold for the word with the highest contribution). Gender revealing words are masked.}
  \label{fig:first}
\end{figure}

The prevalence of accuracy as the single metric to optimize in most popular benchmarks \cite{torralba2011unbiased} has made other aspects of the models, such as fairness, cost, or efficiency, not a priority (and thus, something to not look into). But societal bias is a transversal problem that affects a variety of tasks within computer vision, such as \textit{facial recognition}, with black women having higher error rates than white men \cite{buolamwini2018gender}; \textit{object classification}, with kitchen objects being associated with women with higher probabilities than with men \cite{zhao2017men}; or \textit{pedestrian detection}, with lighter skin individuals showing higher detection rates than darker skin people \cite{wilson2019predictive}. Although the causes of societal bias in different computer vision systems may be similar% (e.g., use of biased datasets, use of accuracy as the only metric to optimise, etc.)
, the consequences are particular and require specific solutions for each task.

We examine and quantify societal bias in image captioning (Figure \ref{fig:first}).  Image captioning has achieved state-of-the-art accuracy on MSCOCO captions dataset \cite{chen2015microsoft} by means of pre-trained visual and language Transformers \cite{li2020oscar}. By leveraging very large-scale collections of data (\eg, Google Conceptual Captions \cite{sharma2018conceptual} with about 3.3 million image-caption pairs crawled from the Internet), self-attention-based models \cite{vaswani2017attention} have the potential to learn world representations according to the training distribution. However, these large amounts of data, often without (or with minimal) curation, conceal multiple problems, including the normalization of abuse or the encoding of discrimination \cite{excavating,birhane2021multimodal,prabhu2020large}. So, once image captioning models have achieved outstanding performance on evaluation benchmarks, a question arises: are these models safe and fair to everyone?

We are not the first to formulate this question.  Image captioning has been shown to reproduce gender \cite{burns2018women} and racial \cite{zhao2021captionbias} bias. By demonstrating the existence of societal bias in image captioning, the pioneering work in \cite{burns2018women} set the indispensable seed to continue to investigate this problem, which we believe is far from being solved. We argue that one of the aspects that remains open is the quantification and evaluation of societal bias in image captioning. So far, a variety of metrics have been applied to assess different aspects of societal bias in human and model-generated captions, such as whether the representation of different subgroups is balanced \cite{burns2018women,zhao2021captionbias} or whether the protected attributes\footnote{\textit{Protected attribute} refers to a demographic variable (age, gender, race, \etc) that a model should not use to produce an output.} values (\eg, \textit{female}, \textit{male}) are correctly predicted \cite{burns2018women,tang2021mitigating}. However, in Section~\ref{sec:analysis}, we show that current metrics may be insufficient, as they only consider the effects of bias perpetuation to a degree. 

With the aim to identify and correct bias in image captioning, in Section~\ref{sec:ourmetric}, we propose a simple but effective metric that measures not only how much biased a trained captioning model is, but also how much bias is introduced by the model with respect to the training dataset. This simple metric allows us to conduct a comprehensive analysis of image captioning models in terms of gender and racial bias (Section \ref{sec:evaluation}), with an unexpected revelation: the gender equalizer designed to reduce gender bias in \cite{burns2018women} is actually amplifying gender bias when considering the semantics of the whole caption. This discovery highlights, even more, the necessity of a standard, unified metric to measure bias and bias amplification in image captioning, as the efforts to address societal inequalities will be ineffective without a tool to quantify how much bias a system exhibits and where this bias is coming from. We conclude the paper with an analysis of the limitation of the proposed metric in Section~\ref{sec:limitations} and a summary of the main findings in Section~\ref{sec:conclusion}.

% \alert{Another specific challenge for image captioning is to determine where the bias is coming from. Being a multimodal task at the intersection of vision and language, societal bias can be encoded in the image, in the language, or in both. We argue that efforts to address societal bias will be ineffective without a tool to quantify how much bias a system exhibits and where this bias is coming from. \alert{another thing previous work is not looking at is the correlation between bias and accuracy metrics.}

% In this paper we provide a comprehensive overview of the current efforts analysing image captioning (Section 2), we analyse a variety of metrics used before (Section 3), we propose a new metric (Section 4) and we evaluate popular captioning systems in terms of bias (Section 5).}

% \alert{Some summary of the main findings...}

\section{Related work}
\label{sec:relatedwork}

\paragraph{Societal bias in computer vision} 
The problem of bias in large-scale computer vision datasets was first raised by Torralba and Efros in \cite{torralba2011unbiased}, where the differences in the image domain between datasets were explored. Each dataset presented different versions of the same object (\eg, cars in Caltech \cite{fei2004learning} tended to appear sidewise, whereas in ImageNet \cite{deng2009imagenet} were predominantly of racing type), impacting cross-dataset generalization. But it was only recently that societal bias in computer vision was formally investigated. 

In the seminal work of Buolamwini and Gebru \cite{buolamwini2018gender}, commercial face recognition applications were examined across subgroups, demonstrating that performance was different according to the gender and race of each individual, especially misclassifying women with darker skin tones. Similarly, Zhao \etal \cite{zhao2017men} showed not only that images in MSCOCO \cite{lin2014microsoft} were biased towards a certain gender, but also that action and object recognition models amplified such bias in their predictions. With an increased interest in fairness, multiple methods for mitigating the effects of dataset bias have been proposed \cite{zhao2017men,wang2019balanced,wang2019racial,thong2021feature,jia2020mitigating}. 

\vspace{-14pt}
\paragraph{Measuring societal bias} 
Societal bias is a problem with multiple layers of complexity. Even on balanced datasets, models still perpetuate bias \cite{wang2019balanced}, indicating that social stereotypes are occurring at the deepest levels of the image. This makes the manual identification and annotation of biases unfeasible. Thus, the first step towards fighting and mitigating bias is to quantify the problem.

Bias quantification metrics have been introduced for image classification. Zhao \etal \cite{zhao2017men} defined bias based on the co-occurrence of  objects and protected attributes; Wang \etal \cite{wang2019balanced} relied on the accuracy of a classifier when predicting the protected attributed; and Wang and Russakovsky \cite{wang2021directional} extended the definition of bias by including directionality. In addition, REVISE \cite{wang2020revise} and CIFAR-10S \cite{wang2020towards} ease the task of identifying bias on datasets and models, respectively. These solutions, however, cannot be directly applied to image captioning, so specific methods must be developed.

\vspace{-14pt}
\paragraph{Societal bias in image captioning} 
In image captioning \cite{vinyals2015show,xu2015show,you2016image,anderson2018bottom} the input to the model is an image and the output is a natural language sentence. This duality of data modalities makes identifying bias particularly challenging, as it can be encoded in the image and/or in the language. The original work by Burns \etal \cite{burns2018women} showed that captions in MSCOCO \cite{chen2015microsoft} present gender imbalance and proposed an equalizer to force captioning models to produce gender words based on visual evidence. Recently, Zhao \etal \cite{zhao2021captionbias} studied racial bias from several perspectives, including visual representation, sentiment analysis, and semantics. 

Each of these studies, however, uses different evaluation protocols and definitions of bias, lacking of a standard metric. To fill this gap, we propose an evaluation metric to measure not only how biased a model is, but how much it is amplified with respect to the original (biased) dataset. % After extensive experiments, we found that mitigation methods may be actually contributing to bias amplification.

\section{Analysis of fairness metrics}
\label{sec:analysis}
Bias in image captioning has been estimated using different methods: how balanced the prediction of the protected attributed is~\cite{burns2018women}, the overlap of attention maps with segmentation annotations~\cite{burns2018women}, or the difference in accuracy between the different protected attributes~\cite{zhao2021captionbias}. In this section, we thoroughly examine existing fairness evaluation metrics and their shortcomings when applied to image captioning.

\vspace{-14pt}
\paragraph{Notation}
Let $\mathcal{D}$ denote the training split of a certain vision dataset with samples $(I, y, a)$, where $I$ is an image, $y$ is the ground-truth annotation for a certain task, and $a \in \mathcal{A}$ is a protected attribute in set $\mathcal{A}$. The validation/test split is denoted by $\mathcal{D}'$. We assume there is a model $M$ that makes prediction $\hat{y}$ associated with this task from the image, \ie, $\hat{y} = M(I)$. For image captioning, we define a ground-truth caption $y = (y_1, y_2, \dots, y_n)$ as a sequence of $n$ tokens.%, and $\mathcal{V}$ as a vocabulary set.

%%%%%%%%%%%%%%%%%%%%%%%%%%%%%%%%%%%%%%%%%%%%%%%%%%%%%%%%%%%%%%%%%%%%%%%%%%%%%%%%%%%
\subsection{Fairness metrics in image captioning}

\paragraph{Difference in performance} A natural strategy to show bias in image captioning is as the difference in performance between the subgroups of a protected attribute, in terms of accuracy~\cite{burns2018women,tang2021mitigating,zhao2021captionbias}, ratio~\cite{burns2018women}, or sentiment analysis~\cite{zhao2021captionbias}. Quantifying the existence of different behavior according to demographic groups is essential to demonstrate the existence of bias in a model, but it is insufficient for a deeper analysis, as it does not provide information on where the bias comes from, and whether bias is being amplified by the model. Thus, it is good practice to accompany difference in performance with other fairness metrics.

\vspace{-14pt}
\paragraph{Attribute misclassification} Another common metric is to check if the protected attribute has been correctly predicted in the generated caption \cite{burns2018women,tang2021mitigating}.  This assumes that the attribute can be clearly identified in a sentence, which may be the case for some attributes, \eg, age (\textit{a young person}, \textit{a child}) or gender  (\textit{a woman}, \textit{a man}), but not for others, \eg, skin tone. This is critical for two reasons: 1) even when the attribute is not clearly mentioned in a caption, bias can occur through the use of different language to describe different demographic groups; and 2) it only considers the prediction of the protected attribute, ignoring the rest of the sentence which may also exhibit bias.

\vspace{-14pt}
\paragraph{Right for the right reasons} Introduced in \cite{burns2018women}, it measures whether the attention activation maps when generating a protected attribute word $w$ in the caption, \eg, \textit{woman} or \textit{man}, are located in the image region where the evidence about the protected attribute is found, \ie, the person. 
%
% Let $\mathcal{W}$ be a set of words associated with the protected attribute $a$, and $S_I$ be a binary person segmentation map of image $I \in \mathcal{D}'$, so that $S_I(x) = 1$ if pixel $x$ belongs to a person region, and $S_I(x) = 0$ otherwise. For the protected attributed word $w \in \mathcal{W}$ in generated caption $\hat{y} = M(I)$, a visual explanation map $\epsilon_I$ is obtained with Grad-CAM~\cite{selvaraju2017grad}, where $\epsilon_I(x)$ indicates the contribution of $x$ to generate $w$. Then, right for the right reasons is:
% %
% \begin{equation}
%     \text{Right} = {1 \over |\mathcal{D}'|} \sum_{I\in\mathcal{D}'} S_I(\text{argmax}_x \epsilon_I(x)).
% \end{equation}
%
This metric quantifies the important task of whether $w$ is generated based on the person visual evidence or, on the contrary, on the visual context, which has been shown to be one of the sources of bias in image captioning models. However, it has three shortcomings: 1) it needs a shortlist of protected attribute words, and a person segmentation map per image, which may not always be available; 2) it assumes that visual explanations can be generated from the model, which may not always be the case; and 3) it does not consider the potential bias in the rest of the sentence, which (as we show in Section \ref{sec:evaluation}) is another critical source of bias. 

\vspace{-14pt}
\paragraph{Sentence classification} Lastly, Zhao \etal \cite{zhao2021captionbias} introduced the use of sentence classifiers for analyzing racial bias. The reasoning is that if a classifier can distinguish between subgroups in the captions, the captions contain bias. Formally, let $f$ denote a classifier that predicts a protected attribute in $\mathcal{A}$ trained over $\mathcal{D}$, \ie, $\hat{a} = f(y)$, from a caption $y$ in an arbitrary set $\mathcal{H}$ of captions. If the accuracy is higher than the chance rate, $y$ is considered to be biased:
\begin{equation}
    \text{SC} = {1 \over |\mathcal{H}|} \sum_{y \in \mathcal{H}} \mathbb{1}[ f(y) = a ],
\end{equation}
where $\mathbb{1}[\cdot]$ is a indicator function that gives $1$ when the statement provided as the argument holds true and $0$ otherwise. $\mathcal{H}$ typically is the set of all captions generated from the images in the test/validation split $\mathcal{D}'$ of the dataset, \ie, $\mathcal{H} = \{M(I) \mid I \in \mathcal{D}'\}$.

Unlike the previous methods, this metric considers the full context of the caption. However, a major shortcoming is that, when bias exists on the generated data, the contributing source is not identified. Whether the bias comes from the model or from the training data and whether bias is being amplified or not, cannot be concluded. %They don't mask protected words and don't apply noise to the ground truth for comparison. We build on top of this.

% \begin{table*}[t]
% \renewcommand{\arraystretch}{1.1}
% \setlength{\tabcolsep}{4pt}
% \small
% \centering
% \caption{\alert{Summary of fairness metrics.}}
% \begin{tabularx}{0.96\textwidth}{@{}l l c c c c c c@{}}
% \toprule
% & & \multicolumn{3}{c}{Reflects} & & \multicolumn{2}{c}{Requires}  \\ \cline{3-5} \cline{7-8}
% Metric & Target task & Semantics & Context & Amplification & & Attribute Pred. & Segmentation \\
% \midrule
% Difference in performance \cite{burns2018women} & Captioning & - & - & - & & - & -  \\
% Attribute misclassification \cite{burns2018women} & Captioning & - & - & - & & \cmark & - \\
% Right for the right reasons \cite{burns2018women} & Captioning & - & - & - & & \cmark & \cmark   \\
% Sentence classification \cite{zhao2021captionbias} & Captioning & \cmark & \cmark & - & & - & -  \\
% Bias amplification \cite{zhao2017men}  & Classification & - & - & \cmark & & \cmark & -   \\
% Leakage \cite{wang2019balanced} & Classification & - & - & \cmark & & - & -    \\
% \alert{Ours} & Captioning & \cmark & \cmark & \cmark & & - & - \\
% \bottomrule
% \end{tabularx}
% \label{tab:metrics_compare}
% \end{table*}

%%%%%%%%%%%%%%%%%%%%%%%%%%%%%%%%%%%%%%%%%%%%%%%%%%%%%%%%%%%%%%%%%%%%%%%%%%%%%%%%%%%
\subsection{Bias amplification metrics}
There is a family of metrics designed to measure bias amplification on visual recognition tasks. We describe them and analyze the challenges when applied to  captioning. % In Section \ref{sec:ourmetric}, we propose a new metric for bias amplification on image captioning.

\vspace{-14pt}
\paragraph{Bias amplification} 
% Doesn't need labels on the test set
% It doesn't consider bias on the representation (imabalance of protected attributes)
% For captioning, it doesn't consider semantics, synonims "woman cooking", "woman making dinner".
% Doesn't consider content "men like shopping" and "men don't like shopping" will be the same
Proposed in \cite{zhao2017men}, it quantifies the implicit correlations between model prediction $\hat{y} = M(I)$ and the protected attribute $a \in \mathcal{A}$ by means of co-occurrence, and whether these correlations are more prominent in the model predictions or in the training data. Let $\mathcal{L}$ denote the set of possible annotations $l$ in the given task, \ie, $y$ and $\hat{y}$ are in $\mathcal{L}$; $c_{a}$ and $\hat{c}_{a}$ denote the numbers of co-occurrences of $a$ and $l$, counted over $y$ and $\hat{y}$, respectively. Bias is 
\begin{equation}
    \tilde{b}_{al} = \frac{\tilde{c}_{al}}{\sum_{a \in \mathcal{A}} \tilde{c}_{al}},
\end{equation}
where $\tilde{c}$ is either $c$ or $\hat{c}$, and $\tilde{b}$ is either $b$ or $\hat{b}$, respectively.
Then, bias amplification is defined by
\begin{equation}
   \text{BA} = \frac{1}{|\mathcal{L}|}\sum_{a \in \mathcal{A}, l \in \mathcal{L}} (\hat{b}_{al} - b_{al} ) \times \mathbb{1}\left[b_{al} > \frac{1}{|\mathcal{A}|}\right].
   \label{eq:bia}
\end{equation}
% 
%\begin{equation}
%    b(a,t) = p (a = 1 | t = 1), 
%    \label{eq:bia2}
%\end{equation}
%
%\begin{equation}
%    y_{at} = \mathbb{1} [ p (a = 1 | t = 1) > {1 \over {|\mathcal{T}|}} ],
%\end{equation}
%
%with $\hat{b}$ computed in the predicted output, and $b^*$ computed in the training set. 

$\text{BA} > 0$ means that bias is amplified by the model, and otherwise mitigated. This metric is useful for a classification task, such as action or image recognition, for which the co-occurrence can be easily counted. However, one of the major shortcomings is that it ignores that protected attributes may be imbalanced in the dataset, \eg, in MSCOCO images \cite{lin2014microsoft} there are $2.25$ more men than women, which causes most of objects to be correlated with men. To solve this and other issues, Wang and Russakovsky \cite{wang2021directional} proposed an extension called directional bias amplification.

\vspace{-14pt}
\paragraph{Leakage} Another way to quantify bias amplification is leakage \cite{wang2019balanced}, which relies on the existence of a classifier to predict the protected attribute $a$. For a sample $(I, y, a)$ in $\mathcal{D}$ with a ground-truth annotation $y$, a classifier $f$ predicts the attribute $a \in \mathcal{A}$ from either $y$ or $\hat{y} = M(I)$. Using this, the leakage can be formally defined as,
\begin{equation}
\text{Leakage} = \lambda_M - \lambda_D, \label{eq:leakage}
\end{equation}
where
\begin{align}
\lambda_D &= {1 \over {|\mathcal{D}|} }  \sum_{(y, a) \in \mathcal{D}} \mathbb{1} [ f(y) = a ] \label{eq:leakage1}\\
\lambda_M &= {1 \over {|\mathcal{D}|} }  \sum_{(I, a) \in \mathcal{D}} \mathbb{1} [ f(\hat{y}) = a ] \label{eq:leakage2}
\end{align}
A positive leakage indicates that $M$ amplifies the bias with respect to the training data, and mitigates it otherwise. % Similarly to BA, this metric is commonly used for image classification tasks. 

% \paragraph{Demographic parity} \cite{kusner2017counterfactual} assesses the independence between a prediction ˆy and a protected
% attribute v such that p(yˆ=y0
% |v=0)=p(yˆ=y0
% |v=1) 

\vspace{-14pt}
\paragraph{Challenges} The direct application of the above metrics to image captioning presents two major challenges. Let us first assume that, for image captioning, the set of words in the vocabulary %$\mathcal{V}$ 
corresponds to the set $\mathcal{L}$ of annotations in Eq.~(\ref{eq:bia}) under a multi-label setting. The first challenge is that these metrics do not consider the semantics of the words: \eg, in the sentences \textit{a woman is cooking} and \textit{a woman is making dinner}, the tokens \textit{cooking} and \textit{making dinner} would be considered as different annotation $l$. The second challenge is they do not consider the context of each word/task: \eg, the token \textit{cooking} will be seen as the same task in the sentence \textit{a man is cooking} and in \textit{a man is not cooking}.

\section{Bias amplification for image captioning}
\label{sec:ourmetric}
We propose a metric %(shown in \alert{Figure XX}) 
to specifically measure bias amplification in image captioning models, borrowing some ideas from sentence classification \cite{zhao2021captionbias} and leakage \cite{wang2019balanced}. Our metric, named LIC, is built on top of the following hypothesis:

\begin{quote}
\textbf{\textit{Hypothesis 1}.} In an unbiased set of captions, there should not exist differences between how demographic groups are represented.
\end{quote}

% To measure until what extend this affirmation is true, we rely on pre-processing techniques and a sentence classifier.

\paragraph{Caption masking} As discussed in Section \ref{sec:analysis}, for some protected attributes (\eg, age and gender), specific vocabulary may be explicitly used in the captions. For example, consider \textit{gender} as a binary\footnote{Due to the availability of annotations in previous work, in this paper we use the binary simplification of gender, acknowledging that it is not inclusive and should be addressed in future work.} protected attribute $a$ with possible values $\{\textit{female}, \textit{male}\}$. The sentence
\begin{quote}
\centering
A girl is playing piano,
\end{quote}
directly reveals the protected attribute value of the caption, \ie, \textit{female}. To avoid explicit mentions to the protected attribute value, we preprocess captions by masking the words related to that attribute.\footnote{A list of attribute-related words is needed for each protected attribute.} The original sentence is then transformed to the masked sentence
\begin{quote}
\centering
A \colorbox{black}{girl} is playing piano.
\end{quote}
Note that this step is not always necessary, as some protected attributes are not  explicitly revealed in the captions.

\vspace{-14pt}
\paragraph{Caption classification}
We rely on a sentence classifier $f_s$ to estimate societal bias in captions. Specifically, we encode each masked caption $y'$ \footnote{If caption masking is not applied, $y' = y$.} with a natural language encoder $E$ to obtain a sentence embedding $e$, as $e = E(y')$. Then, we input $e$ into the sentence classifier $f_s$, whose aim is to predict the protected attribute $a$ from $y'$ as 
\begin{equation}
    \hat{a} = f_s(E(y'))
\end{equation}
$E$ and $f_s$ are learned on a training split $\mathcal{D}$. According to \textit{Hypothesis 1}, in an unbiased dataset, the classifier $f_s$ should not find enough clues in $y'$ to predict the correct attribute $a$. Thus, $\mathcal{D}$ is considered to be biased if  the empirical probability $p(\hat{a} = a)$ over $\mathcal{D}$ is greater than the chance rate.

\vspace{-14pt}
\paragraph{Bias amplification}
Bias amplification is defined as the bias introduced by a model with respect to the existing bias in the training set. To measure bias amplification, we quantify the difference between the bias in the generated captions set $\hat{\mathcal{D}} = \{\hat{y} = M(I) \mid I \in \mathcal{D}\}$ with respect to the bias in the original captions in the training split  $\mathcal{D}$.

One concern with this definition, particular to image captioning, is the difference in the vocabularies used in the annotations and in the predictions, due to 1) the human generated captions typically come with a richer vocabulary, 2) a model's vocabulary is rather limited, and 3) the vocabulary itself can be biased. Thus, naively applying Eq.~(\ref{eq:leakage}) to image captioning can underestimate bias amplification. To mitigate this problem, we introduce noise into the original human captions until the vocabularies in the two sets (model generated and human generated) are aligned. Formally, let $\mathcal{V}_\text{ann}$ and $\mathcal{V}_\text{pre}$ denote the vocabularies identified for all annotations and predictions in the training set, respectively. For the annotation $y = (y_1, \dots, y_N)$, where $y_n$ is the $n$-th word in $y$, we replace all $y_n$ in  $\mathcal{V}_\text{ann}$ but not in $\mathcal{V}_\text{pre}$ with a special out-of-vocabulary token to obtain perturbed annotations $y^\star$, and we train the classifier $f_s^\star$ over $\{y^\star\}$.

\vspace{-14pt}
\paragraph{The LIC metric} The confidence score $s_a^\star$ is an intermediate result of classifier $f_s^\star$, \ie,
\begin{equation}
 \hat{a} = f_s^\star(y^\star) = \text{argmax}_a s^\star_a(y^\star),   
\end{equation}
and it can be interpreted as a posterior probability $p(\hat{a} = a \mid y^\star)$ of the protected attribute $a$ and can give an extra hint on how much $y^\star$ is biased toward $a$. In order words, not only the successful prediction rate is important to determine the bias, but also how confident the predictions are. The same applies to $\hat{s}_a$ and $\hat{f}_s$ trained with $\{\hat{y}\}$.  We incorporate this information into the Leakage for Image Captioning metric (LIC), through
%
% Following \cite{wang2019balanced}, the LIC (Leakage for Image Captioning) metric is based on the difference of successful attribute prediction rates by $f_s^\star$ and $\hat{f}_s$.
%
% The confidence score $s_a(y)$\footnote{$f(y) = \text{argmax}_a s_a(y)$} of classifier $f \in \{f^\star, \hat{f}\}$, which can be interpreted as a posterior probability $p(a \in \mathcal{A} \mid y)$ of protected attribute $a$, can give an extra hint on how much $y$ is biased toward $a$. We incorporate this into LIC, through
\begin{align}
\text{LIC}_D &= {1 \over {|\mathcal{D}|} }  \sum_{(y^\star, a) \in \mathcal{D}} s^\star_a(y^\star) \mathbb{1} [ f^\star(y^\star) = a ] \label{eq:lic_leakage1}\\
\text{LIC}_M &= {1 \over {|\hat{\mathcal{D}}|} }  \sum_{(\hat{y}, a) \in \hat{\mathcal{D}}} \hat{s}_a(\hat{y}) \mathbb{1} [ \hat{f}(\hat{y}) = a ], \label{eq:lic_leakage2}
\end{align}
so that LIC is finally computed as
\begin{equation}
    \text{LIC} = \text{LIC}_M - \text{LIC}_D.
\end{equation}
where a model is considered to amplify bias if $\text{LIC} > 0$. We refer to  $\hat{s}_a$ as the \textit{bias score}.

\section{Experiments}
\label{sec:evaluation}
% We first empirically validate the LIC metric (Section~\ref{sec:score_exp}). Then, we compare multiple image captioning models in terms of gender bias (Section~\ref{sec:gender_exp}) and racial bias (Section~\ref{sec:race_exp}). Finally, we study where the major contribution of bias is coming from (Section~\ref{sec:where_exp}).

% \vspace{-14pt}
\paragraph{Data} Experiments are conducted on a subset of the MSCOCO captions dataset \cite{chen2015microsoft}. Specifically, we use the images with binary gender and race annotations from \cite{zhao2021captionbias}: \textit{female} and \textit{male} for gender, \textit{darker} and \textit{lighter} skin tone for race.\footnote{Similarly, due to the availability of annotations in previous work, we use a binary simplification for race and skin tone. We acknowledge that these attributes are much more complex in reality.} Annotations are available for images in the validation set with person instances, with a total of $10,780$ images for gender and $10,969$ for race. To train the classifiers, we use a balanced split with equal number of images per protected attribute value, resulting in $5,966$ for training and $662$ for test in gender, and $1,972$ for training and $220$ for test in race. Other metrics are reported on the MSCOCO val set. % This split is used in all the reported metrics.

\vspace{-14pt}
\paragraph{Metrics} We report bias using LIC, together with LIC$_D$ in Eq.~(\ref{eq:lic_leakage1}) and LIC$_M$ in Eq.~(\ref{eq:lic_leakage2}). For gender bias, we also use Ratio~\cite{burns2018women}, Error~\cite{burns2018women}, Bias Amplification (BA)~\cite{zhao2017men}, and Directional Bias Amplification~\cite{wang2021directional}. Directional bias amplification is computed for object $\rightarrow$ gender direction (DBA$_G$) and for gender $\rightarrow$ object direction (DBA$_O$) using MSCOCO objects \cite{lin2014microsoft}. For skin tone, we only use LIC, LIC$_D$, and LIC$_M$, as there are no words we can directly associated with race in the captions to calculate the other metrics. Accuracy is reported in terms of standard metrics BLEU-4~\cite{papineni2002bleu}, CIDEr~\cite{vedantam2015cider}, METEOR~\cite{denkowski2014meteor}, and ROUGE-L~\cite{lin2004rouge}.

\vspace{-14pt}
\paragraph{Models} We study bias on captions generated by the following models: NIC~\cite{vinyals2015show}, SAT~\cite{xu2015show}, FC~\cite{rennie2017self}, Att2in~\cite{rennie2017self}, UpDn~\cite{anderson2018bottom}, Transformer~\cite{vaswani2017attention}, OSCAR~\cite{li2020oscar}, NIC+~\cite{burns2018women}, and NIC+Equalizer~\cite{burns2018women}. NIC, SAT, FC, Att2in, and UpDn are classical CNN~\cite{lecun2015deep} encoder-LSTM~\cite{hochreiter1997long} decoder models. Transformer and OSCAR are Transformer-based~\cite{vaswani2017attention} models, which are the current state-of-the-art in image captioning. NIC+ is a re-implementation of NIC in \cite{burns2018women} trained on the whole MSCOCO and additionally trained on MSCOCO-Bias set consisting of images of male/female. NIC+Equalizer is NIC+ with a gender bias mitigation loss, that forces the model to predict gender words only based on the region of the person. Note that most of the pre-trained captioning models provided by the authors are trained on the Karpathy split \cite{karpathy2015deep}, which uses both train and validation splits for training. As the val set is part of our evaluation, we retrain all the models on the MSCOCO train split only. 

\begin{figure}[t]
  \centering
  \vspace{-5pt}
  \includegraphics[clip, width=0.98\columnwidth]{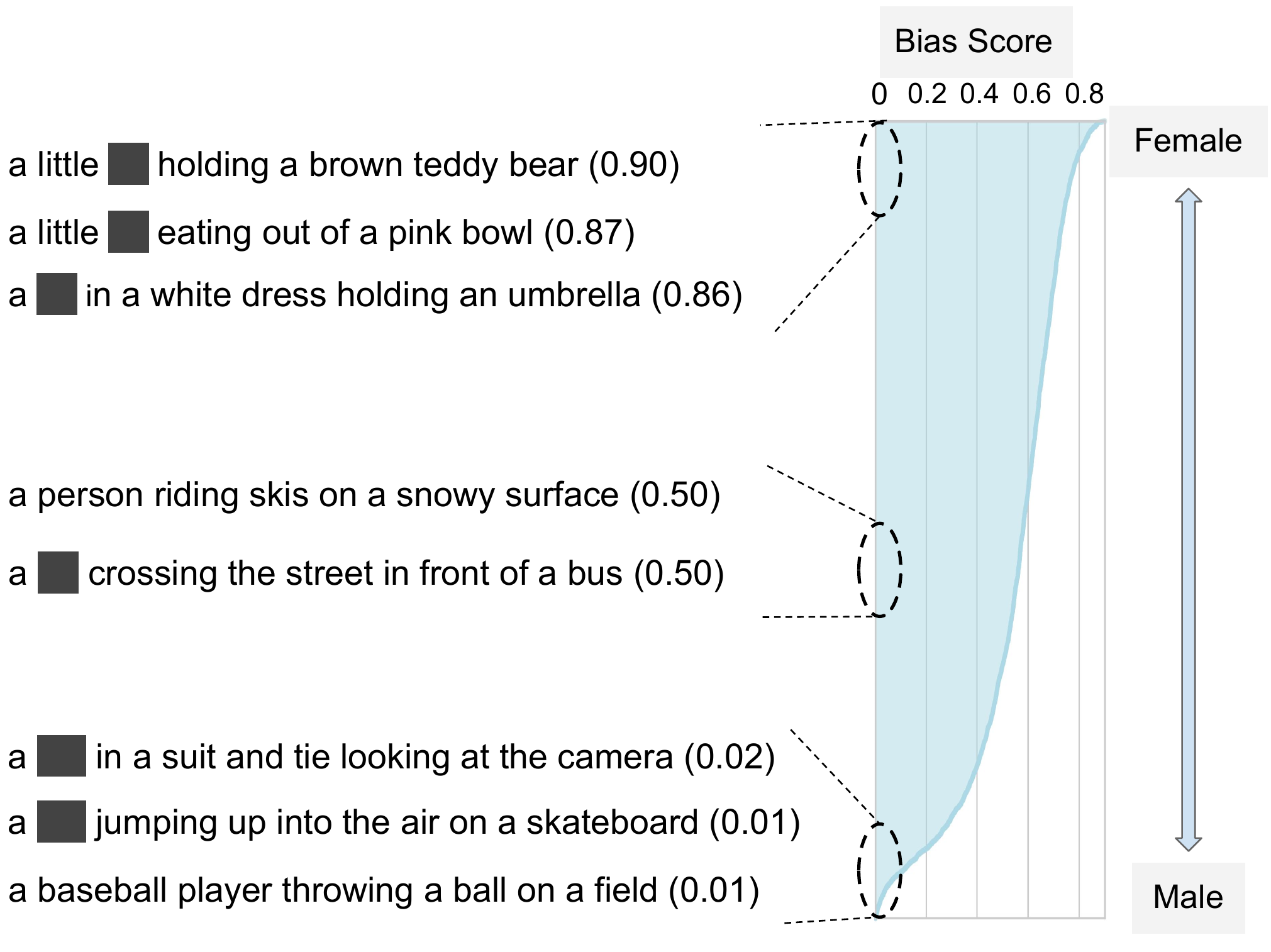}
    \vspace{-5pt}
  \caption{Gender bias score for captions generated with OSCAR. Masked captions are encoded with a LSTM and fed into a gender classifier. Bias score correlates with typical gender stereotypes.}
  \label{fig:score_dist}
\end{figure}

\vspace{-14pt}
\paragraph{LIC metric details} 
For masking, we replace pre-defined gender-related words\footnote{The list of gender-related words can be found in the appendix.} with a special token \texttt{$<$gender$>$}. We do not mask any words for race prediction because race is not commonly explicitly mentioned in the captions. 

The LIC classifier is based on several fully-connected layers on top of a natural language encoder. For the encoder, we use a LSTM \cite{hochreiter1997long} for our main results. We do not initialize the LSTM with pre-computed word embeddings, as they contain bias \cite{bolukbasi2016man,dev2019attenuating}. For completeness, we also report LIC when using BERT \cite{evlin2018bert}, although it has also been shown to exhibit bias \cite{bhardwaj2021investigating,bender2021dangers} and it can affect our metric. BERT is fine-tuned (BERT-ft) or used as is (BERT-pre). The classifier is trained $10$ times with random initializations, and the results are reported by the average and standard deviation. % More details can be found in the appendix.

% The LIC classifier consists of multi-layer perceptron (MLP) with two fully-connected layers and on top of a natural language encoder. For the encoder, we use an LSTM \cite{hochreiter1997long} for our main results. We do not initialize the LSTM with pre-computed word embeddings, as they have been shown to exhibit bias \cite{bolukbasi2016man}. For completeness, we also report LIC when using BERT \cite{evlin2018bert} as encoder, however, it has also been shown to exhibit bias \cite{bhardwaj2021investigating,bender2021dangers} and it can affect our metric. BERT is fine-tuned (BERT-ft) or used as is (BERT-pre).

% Training is conducted on the training set for $20$ epochs for LSTM and BERT-pre, and $5$ epochs for BERT-ft. We use Adam \cite{kingma2014adam} optimizer with a learning rate of $5 \times 10^{-5}$ for LSTM and BERT-pre, and $1 \times 10^{-5}$ for BERT-ft.   

\definecolor{colbest}{rgb}{0.1, 0.6, 0.1}\
\definecolor{colworst}{rgb}{0.75, 0, 0}

\begin{table*}[t]
\renewcommand{\arraystretch}{1.1}
\setlength{\tabcolsep}{5pt}
\small
\centering
\caption{Gender bias and accuracy for several image captioning models. \textcolor{colworst}{Red}/\textcolor{colbest}{green} denotes the worst/best score for each metric. For bias, lower is better. For accuracy, higher is better. BA, DBA$_G$, and DBA$_O$ are scaled by $100$. Unbiased model is LIC$_M = 25$ and $\text{LIC} = 0$.}
\begin{tabularx}{0.93\textwidth}{@{}l r r r r r r r c r r r r@{}}
\toprule
& \multicolumn{7}{c}{Gender bias $\downarrow$} & & \multicolumn{4}{c}{Accuracy $\uparrow$} \\ \cline{2-8} \cline{10-13}
Model & LIC & LIC$_M$ & Ratio & Error & BA & DBA$_G$ & DBA$_O$ & & BLEU-4	& CIDEr & METEOR & ROUGE-L \\
\midrule
NIC \cite{vinyals2015show} &  \textcolor{colbest}{\textbf{3.7}} & \textcolor{colbest}{\textbf{43.2}} & 2.47 & \textcolor{colworst}{\textbf{14.3}} & 4.25 & 3.05 & \textcolor{colbest}{\textbf{0.09}} & & \textcolor{colworst}{\textbf{21.3}} & \textcolor{colworst}{\textbf{64.8}} & \textcolor{colworst}{\textbf{20.7}} & \textcolor{colworst}{\textbf{46.6}}\\
SAT \cite{xu2015show} & 5.1 & 44.4 & 2.06 & 7.3 & 1.14 & 3.53 & 0.15 & & 32.6 & 98.3 & 25.8 & 54.1\\
FC \cite{rennie2017self} & 8.6 & 46.4 & 2.07 & 10.1 & 4.01 & \textcolor{colworst}{\textbf{3.85}} & 0.28 & & 30.5 & 98.0 & 24.7 & 53.5\\
Att2in \cite{rennie2017self} & 7.6 & 45.9 & 2.06 & 4.1 & \textcolor{colbest}{\textbf{0.32}} & 3.60 & \textcolor{colworst}{\textbf{0.29}} & & 33.2 & 105.0 & 26.1 & 55.6\\
UpDn \cite{anderson2018bottom} & 9.0 & 48.0 & 2.15 & 3.7 & 2.78 & 3.61 & 0.28 & & 36.5 & 117.0 & 27.7 & 57.5\\
Transformer \cite{vaswani2017attention} & 8.7 & 48.4 & 2.18 & 3.6 & 1.22 & 3.25 & 0.12 & & 32.3 & 105.3 & 27.0 & 55.1\\
OSCAR \cite{li2020oscar} & 9.2 & 48.5 & 2.06 & \textcolor{colbest}{\textbf{1.4}} & 1.52 & 3.18 & 0.19 & & \textcolor{colbest}{\textbf{40.4}} & \textcolor{colbest}{\textbf{134.0}} & \textcolor{colbest}{\textbf{29.5}} & \textcolor{colbest}{\textbf{59.5}}\\
NIC+ \cite{burns2018women} & 7.2 & 46.7 & \textcolor{colworst}{\textbf{2.89}} & 12.9 & \textcolor{colworst}{\textbf{6.07}} & \textcolor{colbest}{\textbf{2.08}} & 0.17 & & 27.4 & 84.4 & 23.6 & 50.3\\
NIC+Equalizer \cite{burns2018women} & \textcolor{colworst}{\textbf{11.8}} & \textcolor{colworst}{\textbf{51.3}} & \textcolor{colbest}{\textbf{1.91}} & 7.7 & 5.08 & 3.05 & 0.20 & & 27.4 & 83.0 & 23.4	& 50.2\\
\bottomrule
\end{tabularx}
\label{tab:genderbias}
\end{table*}

\begin{figure*}[h!]
\centering
\includegraphics[width = 0.88\textwidth]{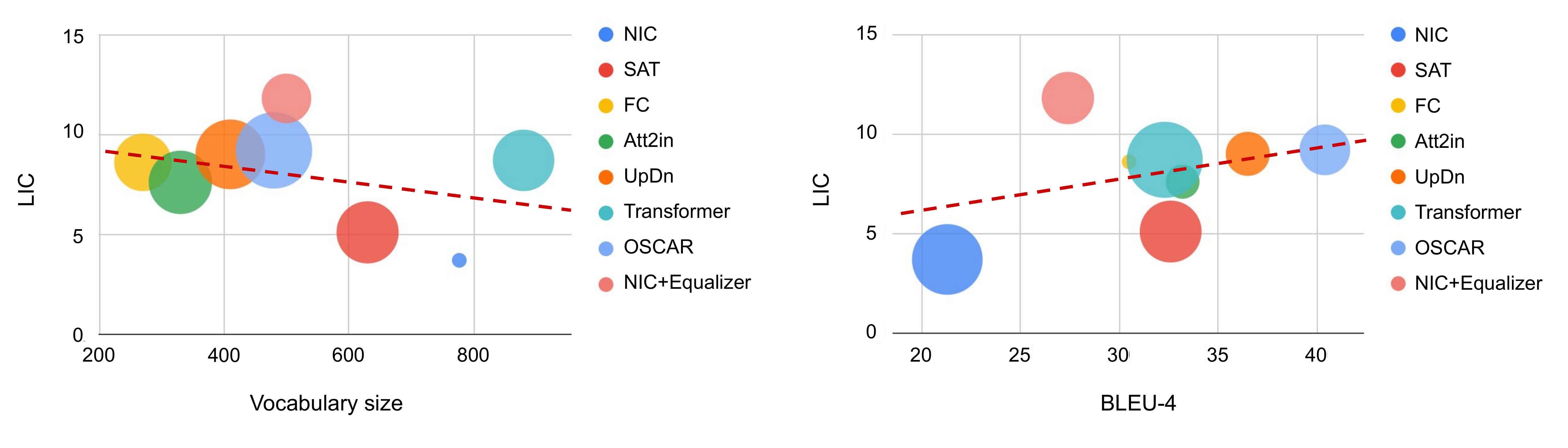}
\vspace{-10pt} 
\caption{LIC vs. Vocabulary size (left) and BLEU-4 score (right). The size of each bubble indicates the BLEU-4 score (left) or the vocabulary size (right). Score tends to decrease with largest vocabularies, but increase with more accurate BLEU-4 models, whereas NIC+Equalizer \cite{burns2018women} is presented as an outlier. The dotted lines indicate the tendency, $R^2=0.153$ (left) and $R^2=0.156$ (right).}
\label{fig:bubble}
\end{figure*}

\begin{table*}[t]
\renewcommand{\arraystretch}{1.1}
\setlength{\tabcolsep}{5pt}
\small
\centering
\caption{Gender bias scores according to LIC, LIC$_M$, and LIC$_D$ for several image captioning models. Captions are encoder with LSTM, BERT-ft, or BERT-pre. Unbiased model is LIC$_M = 25$ and $\text{LIC} = 0$. It shows that LIC is consistent across different language models.}
\begin{tabularx}{0.92\textwidth}{@{}l c c r c c c r c c c r@{}}
\toprule
& \multicolumn{3}{c}{LSTM} & & \multicolumn{3}{c}{BERT-ft} & & \multicolumn{3}{c}{BERT-pre} \\ \cline{2-4} \cline{6-8} \cline{10-12}
Model &  $\text{LIC}_M$ & $\text{LIC}_D$ & LIC & & $\text{LIC}_M$ & $\text{LIC}_D$ & LIC & & $\text{LIC}_M$ & $\text{LIC}_D$ & LIC \\
\midrule
NIC \cite{vinyals2015show} & 43.2 $\pm$ 1.5 & 39.5 $\pm$ 0.9 & \textcolor{colbest}{\textbf{3.7}} & & 47.2 $\pm$ 2.3 & 48.0 $\pm$ 1.2 & \textcolor{colbest}{\textbf{-0.8}} & & 43.2 $\pm$ 1.3 & 41.3 $\pm$ 0.9 & \textcolor{colbest}{\textbf{1.9}}\\
SAT \cite{xu2015show} & 44.4 $\pm$ 1.4 & 39.3 $\pm$ 1.0 & 5.1 & & 48.0 $\pm$ 1.1 & 47.7 $\pm$ 1.4 & 0.3 & & 44.4 $\pm$ 1.5 & 41.5 $\pm$ 0.8 & 2.9 \\
FC \cite{rennie2017self} & 46.4 $\pm$ 1.2 & 37.8 $\pm$ 0.9 & 8.6 & & 48.7 $\pm$ 1.9 &  45.8 $\pm$ 1.3 & 2.9 & & 46.8 $\pm$ 1.4 & 40.4 $\pm$ 0.8 & 6.4\\
Att2in \cite{rennie2017self} & 45.9 $\pm$ 1.1 & 38.3 $\pm$ 1.0 & 7.6 & & 47.8 $\pm$ 2.0 & 46.7 $\pm$ 1.4 & 1.1 & & 45.9 $\pm$ 1.2 & 40.9 $\pm$ 0.9 & 5.0\\
UpDn \cite{anderson2018bottom} & 48.0 $\pm$ 1.3 & 39.0 $\pm$ 0.9 & 9.0 & & 52.0 $\pm$ 1.0 & 47.3 $\pm$ 1.4 & 4.7 & & 48.5 $\pm$ 1.0 & 41.5 $\pm$ 0.9 & 7.0\\
Transformer \cite{vaswani2017attention} & 48.4 $\pm$ 0.8 & 39.7 $\pm$ 0.9 & 8.7 & & 54.1 $\pm$ 1.2 & 48.2 $\pm$ 1.1 & 5.9 & & 47.7 $\pm$ 1.2 & 42.2 $\pm$ 0.9 & 5.5\\
OSCAR \cite{li2020oscar} & 48.5 $\pm$ 1.5 & 39.3 $\pm$ 0.8 & 9.2 & & 52.5 $\pm$ 1.8 & 47.6 $\pm$ 1.2 & 4.9 & & 48.1 $\pm$ 1.1 &41.1 $\pm$ 0.9 & 7.0\\
NIC+ \cite{burns2018women} & 46.7 $\pm$ 1.2 & 39.5 $\pm$ 0.6 & 7.2 & & 49.5 $\pm$ 1.4 &47.7 $\pm$ 1.5 & 1.8 & & 46.4 $\pm$ 1.2 & 41.0 $\pm$ 0.9 & 5.4 \\
NIC+Equalizer \cite{burns2018women}   & 51.3 $\pm$ 0.7 & 39.5 $\pm$ 0.9 & \textcolor{colworst}{\textbf{11.8}} & & 54.8 $\pm$ 1.1 & 47.5 $\pm$ 1.4 & \textcolor{colworst}{\textbf{7.3}} & & 49.5 $\pm$ 0.7 & 40.9 $\pm$ 0.9 & \textcolor{colworst}{\textbf{8.6}}\\
\bottomrule
\end{tabularx}
\label{tab:lstm-bert}
\end{table*}

\subsection{LIC analysis}
\label{sec:score_exp}
We qualitatively analyze the LIC metric to verify whether it is consistent with human intuition. We generate captions in the test set with OSCAR, mask the gender-related words, and encode the masked captions with a LSTM classifier to compute LIC bias score, $\hat{s}_a$, for the gender attribute, as formulated in Section~\ref{sec:ourmetric}. Then, we manually inspect the captions and their associated bias score.

Figure~\ref{fig:score_dist} shows generated captions with higher, middle, and  lower bias scores. The bias score assigned to each caption matches typical gender stereotypes. For example, the third caption from the top, ``a \colorbox{black}{aa} in a white dress holding an umbrella'', yields a very high bias score for \textit{female}, probably due to the stereotype that the people who wear dresses and holds umbrellas tend to be women. On the contrary, the bottom caption, ``a baseball player throwing a ball on a field'', with one of the lowest scores assigned to \textit{female}, perpetuates the stereotype that baseball players are mostly men.
Additionally, when inspecting the captions with a bias score around 0.5, we see that the descriptions tend to be more neutral and without strong gender stereotypes. This support the importance of including $s_a^\star$ and $\hat{s}_a$ in the LIC computation, as in Eqs.~(\ref{eq:lic_leakage1}) and (\ref{eq:lic_leakage2}).

\subsection{Quantification of gender bias}
\label{sec:gender_exp}
We evaluate the gender bias of different captioning models in terms of LIC together with adaptations of existing bias metrics. For BA, we use the top $1,000$ common words in the captions as $\mathcal{L}$, whereas for $\text{DBA}_G$ and $\text{DBA}_O$, we use MSCOCO objects \cite{lin2014microsoft}. More details can be found in the appendix. Results are shown in Table~\ref{tab:genderbias}. We also show the relationship between the quality of a caption, in terms of vocabulary and BLEU-4 score, with LIC in Figure~\ref{fig:bubble}. Finally, we compare LIC when using different language encoders in Table~\ref{tab:lstm-bert}. The main observations are summarized below.

\textbf{\emph{Observation 1.1.} All the models amplify gender bias.} In Table~\ref{tab:genderbias}, all the models have a LIC$_M$ score well over the unbiased model (LIC$_M = 25$), with the lowest score being $43.2$ for NIC. When looking at LIC, which indicates how much bias is introduced by the model with respect to the human captions, also all the models exhibit bias amplification, again with NIC having the lowest score. NIC is also the model that performs the worst in terms of accuracy, which provides some hints about the relationship between accuracy and bias amplification (\textit{Observation 1.4}).

\textbf{\emph{Observation 1.2.} Bias metrics are not consistent.} As analyzed in Section~\ref{sec:analysis}, different metrics measure different aspects of the bias, so it is expected to produce different results, which may lead to different conclusions. Nevertheless, all the models show bias in all the metrics except Ratio (Table~\ref{tab:genderbias}). However, the relationship between the bias and the models presents different tendencies. For instance, NIC+Equalizer shows the largest bias in LIC (\textit{Observation 1.3}) while Att2in has the largest bias in $\text{DBA}_O$. % Overall, the Transformer-based models (Transformer, OSCAR) exhibit less bias than the CNN+LSTM-based models (NIC, SAT, FC, Att2in, UpDn), which also correlates with the performance on accuracy metrics (\textit{Observation 4}).

\textbf{\emph{Observation 1.3.} NIC+Equalizer increases gender bias with respect to the baseline.}
One of the most surprising findings is that even though NIC+Equalizer successfully mitigates gender misclassification when compared to the baseline NIC+ (Error: $12.9 \rightarrow 7.7$ in Table~\ref{tab:genderbias}), it actually increases gender amplification bias by $+4.6$ in LIC. This unwanted side-effect may be produced by the efforts of predicting gender correctly according to the image. As shown in Figure~\ref{fig:first}, NIC+Equalizer tends to produce words that, conversely, are strongly associated with that gender, even if they are not in the image. Results on $\text{DBA}_O$ support this reasoning, revealing that given a gender, NIC+Equalizer rather produces words correlated with that gender. 

\textbf{\emph{Observation 1.4.} LIC tends to increase with BLEU-4, and decrease with vocabulary size.}
Figure~\ref{fig:bubble} shows that larger the vocabulary, the lower the LIC score. This implies that the variety of the words used in the captions is important to suppress gender bias.
As per accuracy, we find that the higher the BLEU-4, the larger the bias tends to be. In other words, even though better models produce better captions, they rely on encoded clues that can identify gender.
% Thus, we can conclude that evaluating captioning models by accuracy only runs the risk of amplifying gender bias.

\textbf{\emph{Observation 1.5.} LIC is robust against encoders.}
In Table~\ref{tab:lstm-bert}, we explore how the selection of language models affects the results of LIC, LIC$_M$, and LIC$_D$ when using LSTM, BERT-ft, and BERT-pre encoders. Although BERT is known to contain gender bias itself, the tendency is maintained within the three language models: NIC shows the least bias, whereas NIC+Equalizer shows the most.

\subsection{Quantification of racial bias}
\label{sec:race_exp}
Results for racial bias when using LSTM as encoder are reported in Table~\ref{tab:race}, leading to the following observations.

\textbf{\emph{Observation 2.1.} All the models amplify racial bias.} 
As with gender, all models exhibits $\text{LIC}>0$.  The magnitude difference of racial bias between the models is smaller than in the case of gender (the standard deviation of LIC among the models is $2.4$ for gender and $1.3$ for race). This indicates that racial bias is amplified without much regard to the structure or performance of the model. In other words, as all the models exhibit similar tendencies of bias amplification, the problem may not only be on the model structure itself but on how image captioning models are trained.

\textbf{\emph{Observation 2.2.} Racial bias is not as apparent as gender bias.}
$\text{LIC}_M$ scores in Table~\ref{tab:race} are consistently smaller than in Table~\ref{tab:lstm-bert}. The mean of the $\text{LIC}_M$ score of all the models is $47.0$ for gender and $33.7$ for race, which is closer to the random chance. 

\textbf{\emph{Observation 2.3.} NIC+Equalizer does not increase racial bias with respect to the baseline.}  
Unlike for gender bias, NIC+Equalizer does not present more racial bias amplification than NIC+. This indicates that forcing the model to focus on the human area to predict the correct gender does not negatively affect other protected attributes. % Thus, it may imply that using the prediction of the protected attributes may involve bias amplification of the protected attributes.

\begin{table}[t]
\renewcommand{\arraystretch}{1.1}
\setlength{\tabcolsep}{7pt}
\small
\centering
\caption{Racial bias scores according to LIC, LIC$_M$, and LIC$_D$. Captions are not masked and are encoder with LSTM.}
\begin{tabularx}{0.42\textwidth}{@{}l c c r@{}}
\toprule
Model &  LIC$_M$ & LIC$_D$ & LIC \\
\midrule
NIC \cite{vinyals2015show} & 33.3 $\pm$ 1.9 & 27.6 $\pm$ 1.0 & 5.7\\
SAT \cite{xu2015show} & 31.3 $\pm$ 2.3 & 26.8 $\pm$ 0.9 & \textcolor{colbest}{\textbf{4.5}} \\
FC \cite{rennie2017self} & 33.6 $\pm$ 1.0 & 26.0 $\pm$ 0.8 & 7.6  \\
Att2in \cite{rennie2017self} & 35.2 $\pm$ 2.3 & 26.6 $\pm$ 0.9 & \textcolor{colworst}{\textbf{8.6}} \\
UpDn \cite{anderson2018bottom} & 34.4 $\pm$ 2.1 & 26.6 $\pm$ 0.9 & 7.8 \\
Transformer \cite{vaswani2017attention} & 33.3 $\pm$ 2.3 & 27.2 $\pm$ 0.8 & 6.1 \\
OSCAR \cite{li2020oscar} & 32.9 $\pm$ 1.8 & 27.0 $\pm$ 1.0 & 5.9 \\
NIC+ \cite{burns2018women} & 34.9 $\pm$ 1.5 & 27.3 $\pm$ 1.2 & 7.6 \\
NIC+Equalizer \cite{burns2018women} & 34.5 $\pm$ 2.8 & 27.3 $\pm$ 0.8 & 7.2 \\
\bottomrule
\end{tabularx}
\label{tab:race}
\end{table}

\subsection{Visual and language contribution to the bias}
\label{sec:where_exp}
As image captioning is a multimodal task involving visual and language information, bias can be introduced by the image, the language, or both. Next, we investigate which modality contributes the most to gender bias by analyzing the behavior when using partially masked images. 

We  define three potential sources of bias: 1) the objects being correlated with the gender \cite{zhao2017men,wang2019balanced,wang2021directional}, 2) the gender of the person in the image \cite{burns2018women}, and 3) the language model itself \cite{bender2021dangers,bolukbasi2016man}. To examine them, we mask different parts of the image accordingly: 1) the object that exhibits the highest correlation with gender according to the BA metric, 2) the person, 3) both of the correlated objects and the person. We analyze SAT \cite{xu2015show} and OSCAR \cite{li2020oscar} as representative models of classical and state-of-the-art captioning, respectively. The details of the experiments can be found in the appendix.
$\text{LIC}_M$ scores are shown in Table~\ref{tab:source}. 

\textbf{\emph{Observation 3.1.} The contribution of objects to gender bias is minimal.}
Results \textit{w/o object} show that masking objects do not considerably mitigate gender bias in the generated captions. Compared to the original $\text{LIC}_M$, the score decreases only $-1.5$ for SAT, and $-2.3$ for OSCAR, concluding that objects in the image have little impact to the gender bias in the final caption.

\textbf{\emph{Observation 3.2.} The contribution of people to gender bias is higher than objects.}
Results \textit{w/o person} show that by masking people in the images, we can reduce bias significantly compared to when hiding objects, indicating that regions associated with humans are the primary source of gender bias among the contents in the image.

\begin{figure}[h]
\centering
\includegraphics[width = 0.98\columnwidth]{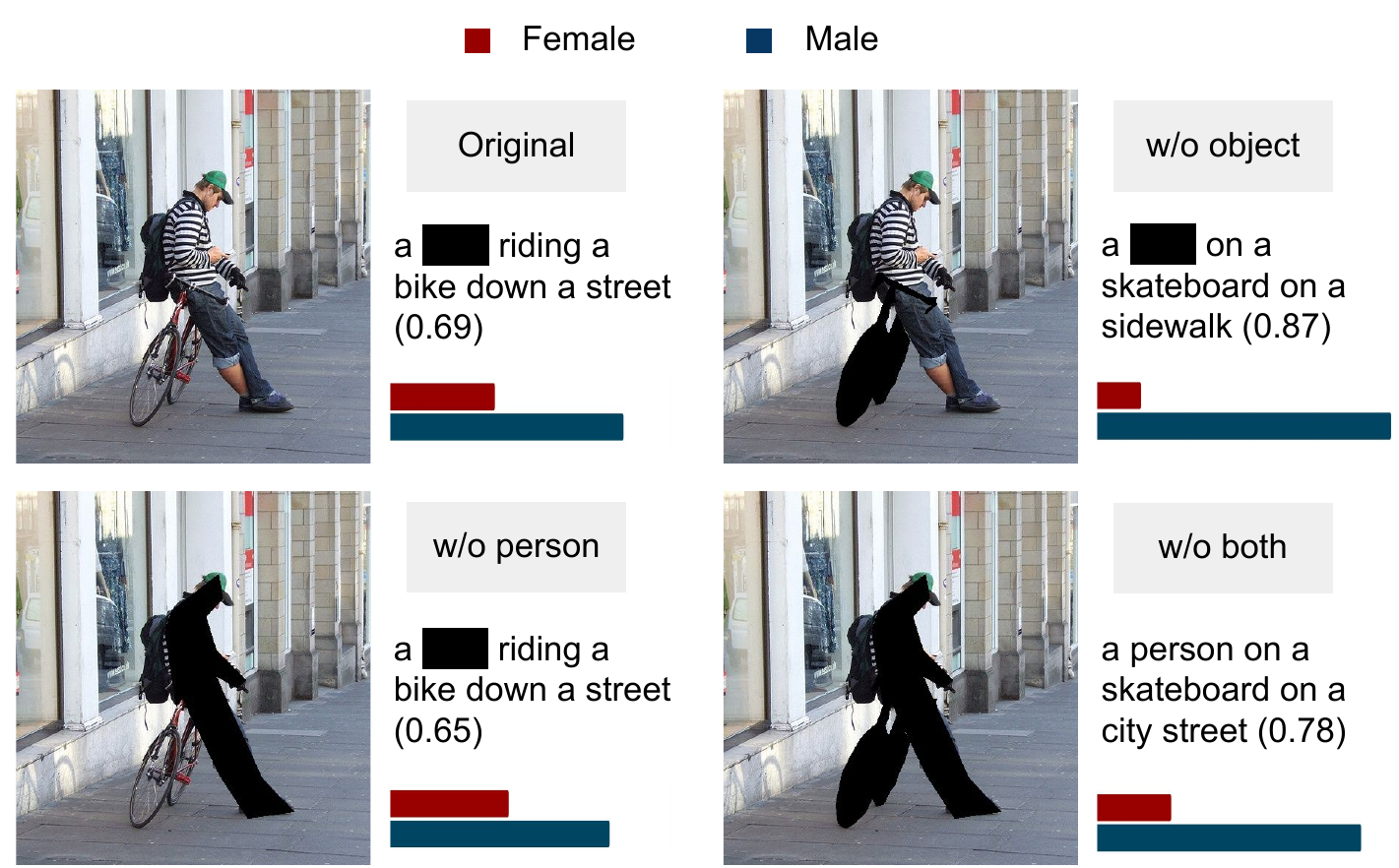}
\vspace{-3pt} 
\caption{Generated captions and bias scores when images are partly masked. The bias score does not decrease when the object (bicycle) and the person (man) are masked.}
\label{fig:masks}
\end{figure}

\textbf{\emph{Observation 3.3.} Language models are a major source of gender bias.}
Results \textit{w/o both} show that even when the gender-correlated objects and people are removed from the images, the generated captions have a large bias ($\Delta_\text{Unbias}$ is $+12.2$ for SAT, $+14.0$ for OSCAR). This indicates that the language model itself is producing a large portion of the bias. To reduce it, it may not be enough to only focus on the visual content, but efforts should also be focused on the language model. Figure~\ref{fig:masks} shows the generated captions and their bias score when images are partly masked.

% \begin{table}[t]
% \renewcommand{\arraystretch}{1.1}
% \setlength{\tabcolsep}{4pt}
% \small
% \centering
% \caption{\alert{Diff-rand and Diff-orig denotes that the difference of $\text{LIC}_M$ from that of the non-biased case (25) and the the original $\text{LIC}_M$ (when using the original images), respectively.}}
% \begin{tabularx}{0.98\columnwidth}{@{}l c c c @{}}
% \toprule
% & \multicolumn{3}{c}{$\text{LIC}_M$} \\ \cline{2-4}
% Model &  MO & MP & MB\\
% \midrule
% SAT \cite{xu2015show} & 40.3 $\pm$ 1.5  & 36.7 $\pm$ 1.3 & 35.0 $\pm$ 0.8\\
% \hspace{5pt} Diff-rand $\downarrow$ (25) & \textcolor{colworst}{\textbf{15.3}} & 11.7 & \textcolor{colbest}{\textbf{10.0}} \\
% \hspace{5pt} Diff-orig $\uparrow$ (41.8 $\pm$ 1.3)& \textcolor{colworst}{\textbf{1.5}} & 5.1 & \textcolor{colbest}{\textbf{6.8}} \\
% OSCAR \cite{li2020oscar} & 43.1 $\pm$ 1.2 & 38.8 $\pm$ 1.3 & 37.3 $\pm$ 1.5 \\
% \hspace{5pt} Diff-rand $\downarrow$ (25) & \textcolor{colworst}{\textbf{18.1}} & 13.8 & \textcolor{colbest}{\textbf{12.3}} \\
% \hspace{5pt} Diff-orig $\uparrow$ (45.6 $\pm$ 1.4) & \textcolor{colworst}{\textbf{2.5}} & 6.8 & \textcolor{colbest}{\textbf{8.3}} \\
% \bottomrule
% \end{tabularx}
% \label{tab:source}
% \end{table}

\begin{table}[t]
\renewcommand{\arraystretch}{1.1}
\setlength{\tabcolsep}{6pt}
\small
\centering
\caption{Gender bias results with partially masked images. $\Delta_\text{Unbias}$ shows the difference with respect to a non-biased model (LIC$_M = 25.0$), and $\Delta_\text{Original}$ with respect to the non-masked case.}
\begin{tabularx}{0.98\columnwidth}{@{}l l c r r @{}}
\toprule
Model & Image & $\text{LIC}_M$ & $\Delta_\text{Unbias}$ & $\Delta_\text{Original}$\\
\midrule
SAT \cite{xu2015show} & Original & $44.4 \pm 1.4$ & $+19.4$ & $0.0$\\
 & w/o object &  $42.9 \pm 1.6$ & $+17.9$ & $-1.5$ \\
 & w/o person & $39.1 \pm 1.4$ & $+14.1$ & $-5.3$\\
 & w/o both & $37.2 \pm 0.8$ & $+12.2$ & $-7.2$\\
\midrule
OSCAR \cite{li2020oscar} & Original & $48.5 \pm 1.5$ & $+23.2$ & $0.0$\\
 & w/o object & $46.2 \pm 1.3$ & $+21.2$ & $-2.3$\\
 & w/o person & $39.7 \pm 1.3$ & $+14.7$ & $-8.8$\\
 & w/o both & $39.0 \pm 1.5$ & $+14.0$ & $-9.5$\\
\bottomrule
\end{tabularx}
\label{tab:source}
\end{table}

\section{Limitations}
\label{sec:limitations}
In Section~\ref{sec:analysis}, we analyzed multiple fairness metrics and their limitations when applied to image captioning. We proposed LIC with the aim to overcome these limitations and unify the evaluation of societal bias in image captioning. However, LIC also presents several limitations.

\vspace{-14pt}
\paragraph{Annotations} LIC needs images to be annotated with their protected attribute. Annotations are not only costly, but may also be problematic. For example, the classification of race is controversial and strongly associated with the cultural values of each annotator~\cite{khan2021one}, whereas gender is commonly classified as a binary $\{\textit{female}, \textit{male}\}$ attribute, lacking inclusiveness with non-binary and other-gender realities.

\vspace{-14pt}
\paragraph{Training} A classifier needs to be trained to make predictions about the protected attributes. The initialization of the model and the amount of training data may impact on the final results. To mitigate this stochastic effect, we recommended to report results conducted on multiple runs.

\vspace{-14pt}
\paragraph{Pre-existing bias} The language encoder may propagate extra bias into the metric if using pretrained biased models, \eg, word embeddings or BERT. To avoid this, we recommend as much random weight initialization as possible.

\section{Conclusion}
\label{sec:conclusion}
This paper proposed LIC, a metric to quantify societal bias amplification in image captioning. LIC is built on top of the idea that there should not be differences between how demographic subgroups are described in captions. The existence of a classifier that predicts gender and skin tone from generated captions more accurately than from human captions, indicated that image captioning models amplify gender and racial bias. Surprisingly, the gender equalizer designed for bias mitigation presented the highest gender bias amplification, highlighting the need of a bias amplification metric for image captioning.

\vspace{-14pt}
\paragraph{Acknowledgements}
Work partly supported by JST CREST Grant No.~JPMJCR20D3, JST FOREST Grant No.~JPMJFR216O, and JSPS KAKENHI, Japan.

%%%%%%%%% REFERENCES
{\small
\bibliographystyle{ieee_fullname}
\bibliography{egbib}

\begin{thebibliography}{10}\itemsep=-1pt

\bibitem{alvi2018turning}
Mohsan Alvi, Andrew Zisserman, and Christoffer Nell{\aa}ker.
\newblock Turning a blind eye: Explicit removal of biases and variation from
  deep neural network embeddings.
\newblock In {\em ECCV Workshops}, 2018.

\bibitem{anderson2018bottom}
Peter Anderson, Xiaodong He, Chris Buehler, Damien Teney, Mark Johnson, Stephen
  Gould, and Lei Zhang.
\newblock Bottom-up and top-down attention for image captioning and visual
  question answering.
\newblock In {\em CVPR}, 2018.

\bibitem{bender2021dangers}
Emily~M Bender, Timnit Gebru, Angelina McMillan-Major, and Shmargaret
  Shmitchell.
\newblock On the dangers of stochastic parrots: Can language models be too big?
\newblock In {\em ACM FAccT}, 2021.

\bibitem{bhardwaj2021investigating}
Rishabh Bhardwaj, Navonil Majumder, and Soujanya Poria.
\newblock Investigating gender bias in {BERT}.
\newblock {\em Cognitive Computation}, 2021.

\bibitem{birhane2021multimodal}
Abeba Birhane, Vinay~Uday Prabhu, and Emmanuel Kahembwe.
\newblock Multimodal datasets: Misogyny, pornography, and malignant
  stereotypes.
\newblock {\em arXiv preprint arXiv:2110.01963}, 2021.

\bibitem{bolukbasi2016man}
Tolga Bolukbasi, Kai-Wei Chang, James~Y Zou, Venkatesh Saligrama, and Adam~T
  Kalai.
\newblock Man is to computer programmer as woman is to homemaker? debiasing
  word embeddings.
\newblock {\em NeurIPS}, 2016.

\bibitem{buolamwini2018gender}
Joy Buolamwini and Timnit Gebru.
\newblock Gender shades: Intersectional accuracy disparities in commercial
  gender classification.
\newblock In {\em ACM FAccT}, 2018.

\bibitem{burns2018women}
Kaylee Burns, Lisa~Anne Hendricks, Kate Saenko, Trevor Darrell, and Anna
  Rohrbach.
\newblock Women also snowboard: Overcoming bias in captioning models.
\newblock In {\em ECCV}, 2018.

\bibitem{chen2015microsoft}
Xinlei Chen, Hao Fang, Tsung-Yi Lin, Ramakrishna Vedantam, Saurabh Gupta, Piotr
  Doll{\'a}r, and C~Lawrence Zitnick.
\newblock Microsoft {COCO} captions: Data collection and evaluation server.
\newblock {\em arXiv preprint arXiv:1504.00325}, 2015.

\bibitem{excavating}
Kate Crawford and Trevor Paglen.
\newblock Excavating {AI}: The politics of training sets for machine learning.
\newblock \url{https://excavating.ai}, 2019.
\newblock Accessed: 2021-11-12.

\bibitem{de2019does}
Terrance de Vries, Ishan Misra, Changhan Wang, and Laurens van~der Maaten.
\newblock Does object recognition work for everyone?
\newblock In {\em CVPR Workshops}, 2019.

\bibitem{deng2009imagenet}
Jia Deng, Wei Dong, Richard Socher, Li-Jia Li, Kai Li, and Li Fei-Fei.
\newblock Image{N}et: A large-scale hierarchical image database.
\newblock In {\em CVPR}, 2009.

\bibitem{denkowski2014meteor}
Michael Denkowski and Alon Lavie.
\newblock Meteor universal: Language specific translation evaluation for any
  target language.
\newblock In {\em Workshop on statistical machine translation}, 2014.

\bibitem{dev2019attenuating}
Sunipa Dev and Jeff Phillips.
\newblock Attenuating bias in word vectors.
\newblock In {\em International Conference on Artificial Intelligence and
  Statistics}. PMLR, 2019.

\bibitem{evlin2018bert}
Jacob Devlin, Ming{-}Wei Chang, Kenton Lee, and Kristina Toutanova.
\newblock {BERT:} pre-training of deep bidirectional transformers for language
  understanding.
\newblock In {\em {NAACL-HLT} {(1)}}, 2019.

\bibitem{d2020data}
Catherine D'ignazio and Lauren~F Klein.
\newblock {\em Data feminism}.
\newblock MIT press, 2020.

\bibitem{fei2004learning}
Li Fei-Fei, Rob Fergus, and Pietro Perona.
\newblock Learning generative visual models from few training examples: An
  incremental bayesian approach tested on 101 object categories.
\newblock In {\em CVPR Workshops}, 2004.

\bibitem{hochreiter1997long}
Sepp Hochreiter and J{\"u}rgen Schmidhuber.
\newblock Long short-term memory.
\newblock {\em Neural computation}, 9(8), 1997.

\bibitem{jia2020mitigating}
Shengyu Jia, Tao Meng, Jieyu Zhao, and Kai-Wei Chang.
\newblock Mitigating gender bias amplification in distribution by posterior
  regularization.
\newblock {\em ACL}, 2020.

\bibitem{karpathy2015deep}
Andrej Karpathy and Li Fei-Fei.
\newblock Deep visual-semantic alignments for generating image descriptions.
\newblock In {\em CVPR}, 2015.

\bibitem{khan2021one}
Zaid Khan and Yun Fu.
\newblock One label, one billion faces: Usage and consistency of racial
  categories in computer vision.
\newblock In {\em ACM FAccT}, 2021.

\bibitem{lecun2015deep}
Yann LeCun, Yoshua Bengio, and Geoffrey Hinton.
\newblock Deep learning.
\newblock {\em Nature}, 521(7553), 2015.

\bibitem{li2020oscar}
Xiujun Li, Xi Yin, Chunyuan Li, Pengchuan Zhang, Xiaowei Hu, Lei Zhang, Lijuan
  Wang, Houdong Hu, Li Dong, Furu Wei, et~al.
\newblock Oscar: Object-semantics aligned pre-training for vision-language
  tasks.
\newblock In {\em ECCV}, 2020.

\bibitem{lin2004rouge}
Chin-Yew Lin.
\newblock Rouge: A package for automatic evaluation of summaries.
\newblock In {\em Text summarization branches out}, 2004.

\bibitem{lin2014microsoft}
Tsung-Yi Lin, Michael Maire, Serge Belongie, James Hays, Pietro Perona, Deva
  Ramanan, Piotr Doll{\'a}r, and C~Lawrence Zitnick.
\newblock Microsoft {COCO}: Common objects in context.
\newblock In {\em ECCV}, 2014.

\bibitem{papineni2002bleu}
Kishore Papineni, Salim Roukos, Todd Ward, and Wei-Jing Zhu.
\newblock {BLEU}: a method for automatic evaluation of machine translation.
\newblock In {\em ACL}, 2002.

\bibitem{prabhu2020large}
Vinay~Uday Prabhu and Abeba Birhane.
\newblock Large image datasets: A pyrrhic win for computer vision?
\newblock {\em arXiv preprint arXiv:2006.16923}, 2020.

\bibitem{rennie2017self}
Steven~J Rennie, Etienne Marcheret, Youssef Mroueh, Jerret Ross, and Vaibhava
  Goel.
\newblock Self-critical sequence training for image captioning.
\newblock In {\em CVPR}, 2017.

\bibitem{sharma2018conceptual}
Piyush Sharma, Nan Ding, Sebastian Goodman, and Radu Soricut.
\newblock Conceptual captions: A cleaned, hypernymed, image alt-text dataset
  for automatic image captioning.
\newblock In {\em ACL}, 2018.

\bibitem{stock2018convnets}
Pierre Stock and Moustapha Cisse.
\newblock Conv{N}ets and {I}mage{N}et beyond accuracy: Understanding mistakes
  and uncovering biases.
\newblock In {\em ECCV}, 2018.

\bibitem{tang2021mitigating}
Ruixiang Tang, Mengnan Du, Yuening Li, Zirui Liu, Na Zou, and Xia Hu.
\newblock Mitigating gender bias in captioning systems.
\newblock In {\em WWW}, 2021.

\bibitem{thong2021feature}
William Thong and Cees~GM Snoek.
\newblock Feature and label embedding spaces matter in addressing image
  classifier bias.
\newblock {\em arXiv preprint arXiv:2110.14336}, 2021.

\bibitem{torralba2011unbiased}
Antonio Torralba and Alexei~A Efros.
\newblock Unbiased look at dataset bias.
\newblock In {\em CVPR}, 2011.

\bibitem{vaswani2017attention}
Ashish Vaswani, Noam Shazeer, Niki Parmar, Jakob Uszkoreit, Llion Jones,
  Aidan~N Gomez, {\L}ukasz Kaiser, and Illia Polosukhin.
\newblock Attention is all you need.
\newblock In {\em NeurIPS}, 2017.

\bibitem{vedantam2015cider}
Ramakrishna Vedantam, C Lawrence~Zitnick, and Devi Parikh.
\newblock {CIDE}r: Consensus-based image description evaluation.
\newblock In {\em CVPR}, 2015.

\bibitem{vinyals2015show}
Oriol Vinyals, Alexander Toshev, Samy Bengio, and Dumitru Erhan.
\newblock Show and tell: A neural image caption generator.
\newblock In {\em CVPR}, 2015.

\bibitem{wang2020revise}
Angelina Wang, Arvind Narayanan, and Olga Russakovsky.
\newblock Revise: A tool for measuring and mitigating bias in visual datasets.
\newblock In {\em ECCV}, 2020.

\bibitem{wang2021directional}
Angelina Wang and Olga Russakovsky.
\newblock Directional bias amplification.
\newblock In {\em ICML}, 2021.

\bibitem{wang2019racial}
Mei Wang, Weihong Deng, Jiani Hu, Xunqiang Tao, and Yaohai Huang.
\newblock Racial faces in the wild: Reducing racial bias by information
  maximization adaptation network.
\newblock In {\em ICCV}, 2019.

\bibitem{wang2019balanced}
Tianlu Wang, Jieyu Zhao, Mark Yatskar, Kai-Wei Chang, and Vicente Ordonez.
\newblock Balanced datasets are not enough: Estimating and mitigating gender
  bias in deep image representations.
\newblock In {\em ICCV}, 2019.

\bibitem{wang2020towards}
Zeyu Wang, Klint Qinami, Ioannis~Christos Karakozis, Kyle Genova, Prem Nair,
  Kenji Hata, and Olga Russakovsky.
\newblock Towards fairness in visual recognition: Effective strategies for bias
  mitigation.
\newblock In {\em CVPR}, 2020.

\bibitem{wilson2019predictive}
Benjamin Wilson, Judy Hoffman, and Jamie Morgenstern.
\newblock Predictive inequity in object detection.
\newblock {\em arXiv preprint arXiv:1902.11097}, 2019.

\bibitem{xu2015show}
Kelvin Xu, Jimmy Ba, Ryan Kiros, Kyunghyun Cho, Aaron Courville, Ruslan
  Salakhudinov, Rich Zemel, and Yoshua Bengio.
\newblock Show, attend and tell: Neural image caption generation with visual
  attention.
\newblock In {\em ICML}, 2015.

\bibitem{yang2020towards}
Kaiyu Yang, Klint Qinami, Li Fei-Fei, Jia Deng, and Olga Russakovsky.
\newblock Towards fairer datasets: Filtering and balancing the distribution of
  the people subtree in the {I}mage{N}et hierarchy.
\newblock In {\em ACM FAccT}, 2020.

\bibitem{you2016image}
Quanzeng You, Hailin Jin, Zhaowen Wang, Chen Fang, and Jiebo Luo.
\newblock Image captioning with semantic attention.
\newblock In {\em CVPR}, 2016.

\bibitem{zhao2021captionbias}
Dora Zhao, Angelina Wang, and Olga Russakovsky.
\newblock Understanding and evaluating racial biases in image captioning.
\newblock In {\em ICCV}, 2021.

\bibitem{zhao2017men}
Jieyu Zhao, Tianlu Wang, Mark Yatskar, Vicente Ordonez, and Kai-Wei Chang.
\newblock Men also like shopping: Reducing gender bias amplification using
  corpus-level constraints.
\newblock In {\em EMNLP}, 2017.

\end{thebibliography}


\begin{thebibliography}{10}\itemsep=-1pt

\bibitem{anderson2018bottom}
Peter Anderson, Xiaodong He, Chris Buehler, Damien Teney, Mark Johnson, Stephen
  Gould, and Lei Zhang.
\newblock Bottom-up and top-down attention for image captioning and visual
  question answering.
\newblock In {\em CVPR}, 2018.

\bibitem{burns2018women}
Kaylee Burns, Lisa~Anne Hendricks, Kate Saenko, Trevor Darrell, and Anna
  Rohrbach.
\newblock Women also snowboard: Overcoming bias in captioning models.
\newblock In {\em ECCV}, 2018.

\bibitem{chen2015microsoft}
Xinlei Chen, Hao Fang, Tsung-Yi Lin, Ramakrishna Vedantam, Saurabh Gupta, Piotr
  Doll{\'a}r, and C~Lawrence Zitnick.
\newblock Microsoft {COCO} captions: Data collection and evaluation server.
\newblock {\em arXiv preprint arXiv:1504.00325}, 2015.

\bibitem{evlin2018bert}
Jacob Devlin, Ming{-}Wei Chang, Kenton Lee, and Kristina Toutanova.
\newblock {BERT:} pre-training of deep bidirectional transformers for language
  understanding.
\newblock In {\em {NAACL-HLT} {(1)}}, 2019.

\bibitem{he2016deep}
Kaiming He, Xiangyu Zhang, Shaoqing Ren, and Jian Sun.
\newblock Deep residual learning for image recognition.
\newblock In {\em CVPR}, pages 770--778, 2016.

\bibitem{hochreiter1997long}
Sepp Hochreiter and J{\"u}rgen Schmidhuber.
\newblock Long short-term memory.
\newblock {\em Neural computation}, 9(8), 1997.

\bibitem{jiang2020defense}
Huaizu Jiang, Ishan Misra, Marcus Rohrbach, Erik Learned-Miller, and Xinlei
  Chen.
\newblock In defense of grid features for visual question answering.
\newblock In {\em CVPR}, pages 10267--10276, 2020.

\bibitem{kingma2014adam}
Diederik~P. Kingma and Jimmy Ba.
\newblock Adam: {A} method for stochastic optimization.
\newblock In {\em ICLR}, 2015.

\bibitem{li2020oscar}
Xiujun Li, Xi Yin, Chunyuan Li, Pengchuan Zhang, Xiaowei Hu, Lei Zhang, Lijuan
  Wang, Houdong Hu, Li Dong, Furu Wei, et~al.
\newblock Oscar: Object-semantics aligned pre-training for vision-language
  tasks.
\newblock In {\em ECCV}, 2020.

\bibitem{lin2014microsoft}
Tsung-Yi Lin, Michael Maire, Serge Belongie, James Hays, Pietro Perona, Deva
  Ramanan, Piotr Doll{\'a}r, and C~Lawrence Zitnick.
\newblock Microsoft {COCO}: Common objects in context.
\newblock In {\em ECCV}, 2014.

\bibitem{ren2016faster}
Shaoqing Ren, Kaiming He, Ross Girshick, and Jian Sun.
\newblock Faster r-cnn: towards real-time object detection with region proposal
  networks.
\newblock {\em IEEE transactions on pattern analysis and machine intelligence},
  2016.

\bibitem{rennie2017self}
Steven~J Rennie, Etienne Marcheret, Youssef Mroueh, Jerret Ross, and Vaibhava
  Goel.
\newblock Self-critical sequence training for image captioning.
\newblock In {\em CVPR}, 2017.

\bibitem{vaswani2017attention}
Ashish Vaswani, Noam Shazeer, Niki Parmar, Jakob Uszkoreit, Llion Jones,
  Aidan~N Gomez, {\L}ukasz Kaiser, and Illia Polosukhin.
\newblock Attention is all you need.
\newblock In {\em NeurIPS}, 2017.

\bibitem{vinyals2015show}
Oriol Vinyals, Alexander Toshev, Samy Bengio, and Dumitru Erhan.
\newblock Show and tell: A neural image caption generator.
\newblock In {\em CVPR}, 2015.

\bibitem{wang2021directional}
Angelina Wang and Olga Russakovsky.
\newblock Directional bias amplification.
\newblock In {\em ICML}, 2021.

\bibitem{xu2015show}
Kelvin Xu, Jimmy Ba, Ryan Kiros, Kyunghyun Cho, Aaron Courville, Ruslan
  Salakhudinov, Rich Zemel, and Yoshua Bengio.
\newblock Show, attend and tell: Neural image caption generation with visual
  attention.
\newblock In {\em ICML}, 2015.

\bibitem{zhao2021captionbias}
Dora Zhao, Angelina Wang, and Olga Russakovsky.
\newblock Understanding and evaluating racial biases in image captioning.
\newblock In {\em ICCV}, 2021.

\bibitem{zhao2017men}
Jieyu Zhao, Tianlu Wang, Mark Yatskar, Vicente Ordonez, and Kai-Wei Chang.
\newblock Men also like shopping: Reducing gender bias amplification using
  corpus-level constraints.
\newblock In {\em EMNLP}, 2017.

\end{thebibliography}
}

\end{document}

\title{Supplementary Material for Quantifying Societal\\Bias Amplification in Image Captioning}

% \and
\maketitle

%%%%%%%%% BODY TEXT
This supplementary material includes:

\begin{itemize}[topsep=0pt,itemsep=-1ex,partopsep=1ex,parsep=1ex]
    \item Experimental details (Appendix~\ref{sec:details}).
    \item List of gender-related words (Appendix~\ref{sec:list-words}).
    \item More visual examples (Appendix~\ref{sec:examples}).
    \item Additional results (Appendix~\ref{sec:moreresults}).
    \item Potential negative impact (Appendix~\ref{sec:impact}).
    %\item Code (external files).
\end{itemize}

\section{Experimental details}
\label{sec:details}
In this section, we provide the details for the experiments.

\subsection{LIC metric training details}
We evaluate three classifiers for LIC (LSTM, BERT-ft, and BERT-pre). Their details and hyperparameters can be found below. 
All the classifiers are trained with Adam~\cite{kingma2014adam}. 
% Training is conducted on the training set for $20$ epochs for LSTM and BERT-pre, and $5$ epochs for BERT-ft. We use Adam \cite{kingma2014adam} optimizer with a learning rate of $5 \times 10^{-5}$ for LSTM and BERT-pre, and $1 \times 10^{-5}$ for BERT-ft.

\begin{itemize}
    \item \textbf{LSTM. } A two-layer bi-directional LSTM~\cite{hochreiter1997long} with a fully-connected layer on top. Weights are initialized randomly and training is conducted on the training set for $20$ epochs, with learning rate $5 \times10^{-5}$.
    \item \textbf{BERT-ft. } BERT-base \cite{evlin2018bert} Transformer with two fully-connected layers with Leaky ReLU activation on top. All the weights are fine-tuned while training. Training is conducted for $5$ epochs with learning rate $1 \times 10^{-5}$. 
    \item \textbf{BERT-pre. } Same architecture as BERT-ft. Only the last fully-connected layers are fine-tuned, whereas BERT weights are frozen. Training is conducted for $20$ epochs with learning rate $5 \times 10^{-5}$. 
\end{itemize}

\subsection{Other metrics details}
Details for computing BA, $\text{DBA}_G$, and $\text{DBA}_O$ metrics.

\begin{itemize}
\item \textbf{BA. } We use nouns, verbs, adjectives, and adverbs of the top $1,000$ common words in the captions as $\mathcal{L}$ and calculate the co-occurrence of the gender words and the common words in the captions. As \cite{zhao2017men}, we filter the words that are not strongly associated with humans by removing words that do not occur with each gender at least $100$ times in the ground-truth captions, leaving a total of $290$ words. 
\item \textbf{$\text{DBA}_G$ and $\text{DBA}_O$. }
Let $p$ denote the probability calculated by the (co-)occurrence. The definition of $\text{DBA}_G$ and $\text{DBA}_O$ \cite{wang2021directional} is:

\begin{equation}
    \text{DBA} = \frac{1}{|\mathcal{L}| |\mathcal{A}|}\sum_{a \in \mathcal{A}, l \in \mathcal{L}} y_{al} \Delta_{al} + (1 - y_{al}) (-\Delta_{al})
\end{equation}
\begin{equation}
    y_{al} = \mathbb{1}\left[p(a,l) > p(a) p(l)\right]
\end{equation}
\begin{equation}
    \Delta_{al} = \left\{
                \begin{array}{ll}
                \hat{p}(a|l) - p(a|l) & \text{for} \ \text{DBA}_G \\
                \hat{p}(l|a) - p(l|a) & \text{for} \  \text{DBA}_O
                \end{array}
                \right.
\end{equation}

For $\text{DBA}_G$, we use the MSCOCO objects \cite{lin2014microsoft} annotated on the images as $\mathcal{L}$ and gender words in the captions as $\mathcal{A}$. 
For $\text{DBA}_O$, we use the MSCOCO objects \cite{lin2014microsoft} in the captions as $\mathcal{L}$ and gender annotations \cite{zhao2021captionbias} as $\mathcal{A}$. 
\end{itemize}

\definecolor{colbest}{rgb}{0.1, 0.6, 0.1}\
\definecolor{colworst}{rgb}{0.75, 0, 0}
 \begin{table*}[t]
 \renewcommand{\arraystretch}{1.1}
 \setlength{\tabcolsep}{4pt}
 \small
 \centering
 \caption{Racial bias scores according to LIC, LIC$_M$, and LIC$_D$ for several image captioning models. Captions are encoder with LSTM, BERT-ft, or BERT-pre. Unbiased model is LIC$_M = 25$ and $\text{LIC} = 0$.}
 \begin{tabularx}{0.87\textwidth}{@{}l c c r c c c r c c c r@{}}
 \toprule
 & \multicolumn{3}{c}{LSTM} & & \multicolumn{3}{c}{BERT-ft} & & \multicolumn{3}{c}{BERT-pre} \\ \cline{2-4} \cline{6-8} \cline{10-12}
 Model &  LIC$_M$ & LIC$_D$ & LIC & & LIC$_M$ & LIC$_D$ & LIC & & LIC$_M$ & LIC$_D$ & LIC \\
 \midrule
 NIC \cite{vinyals2015show} & 33.3 $\pm$ 1.9 & 27.6 $\pm$ 1.0 & 5.7 & & 37.0 $\pm$ 3.0 & 36.7 $\pm$ 1.1 & \textcolor{colbest}{\textbf{0.3}} & & 34.7 $\pm$ 2.1 & 33.6 $\pm$ 1.2 & 1.1\\
 SAT \cite{xu2015show} & 31.3 $\pm$ 2.3 & 26.8 $\pm$ 0.9 & \textcolor{colbest}{\textbf{4.5}} & & 38.1 $\pm$ 2.7 & 36.5 $\pm$ 1.4 & 1.6 & & 33.9 $\pm$ 1.5 & 33.3 $\pm$ 1.3 & \textcolor{colbest}{\textbf{0.6}}\\
 FC \cite{rennie2017self} & 33.6 $\pm$ 1.0 & 26.0 $\pm$ 0.8 & 7.6  & & 40.4 $\pm$ 2.4 & 36.4 $\pm$ 1.6 & 4.0 & & 36.9 $\pm$ 2.2 & 32.6 $\pm$ 1.2 & \textcolor{colworst}{\textbf{4.3}}\\
 Att2in \cite{rennie2017self} & 35.2 $\pm$ 2.3 & 26.6 $\pm$ 0.9 & \textcolor{colworst}{\textbf{8.6}} & & 40.4 $\pm$ 2.0 & 36.1 $\pm$ 1.2 & \textcolor{colworst}{\textbf{4.3}} & & 36.8 $\pm$ 1.9 & 32.7 $\pm$ 1.1 & 4.1\\
 UpDn \cite{anderson2018bottom} & 34.4 $\pm$ 2.1 & 26.6 $\pm$ 0.9 & 7.8 & & 40.2  $\pm$ 1.7 & 36.9 $\pm$ 1.2 & 3.3 & & 36.5 $\pm$ 2.5 & 33.2 $\pm$ 1.2 & 3.3 \\
 Transformer \cite{vaswani2017attention} & 33.3 $\pm$ 2.3 & 27.2 $\pm$ 0.8 &    6.1 & & 39.4 $\pm$ 1.7 & 37.4 $\pm$ 1.3 & 2.0 & & 36.2 $\pm$ 2.2 & 34.1 $\pm$ 1.2 & 2.1 \\
 OSCAR \cite{li2020oscar} & 32.9 $\pm$ 1.8 & 27.0 $\pm$ 1.0 & 5.9 & & 39.4 $\pm$ 2.3 & 36.9 $\pm$ 0.9 & 2.5 & & 35.5 $\pm$ 2.5 & 32.9 $\pm$ 1.1 & 2.6 \\
 NIC+ \cite{burns2018women} & 34.9 $\pm$ 1.5 & 27.3 $\pm$ 1.2 & 7.6 & & 39.5 $\pm$ 2.6 & 37.1 $\pm$ 1.3 & 2.4 & & 36.8 $\pm$ 2.4 & 33.6 $\pm$ 1.3 & 3.2\\
 NIC+Equalizer \cite{burns2018women} & 34.5 $\pm$ 2.8 & 27.3 $\pm$ 0.8 & 7.2 & & 38.7 $\pm$ 3.1 & 36.6 $\pm$ 1.3 & 2.1 & & 36.0 $\pm$ 2.2 & 33.4 $\pm$ 1.4 & 2.6 \\
 \bottomrule
 \end{tabularx}
 \label{tab:race}
 \end{table*}

\subsection{Image masking}
% \subsection{Visual and language contribution to the bias}
Here, we explain how we masked objects and people in the images to estimate the contribution of each modality to the bias.
% In this experiment, the way to mask objects or people in the images is different between the models; SAT \cite{xu2015show} and OSCAR \cite{li2020oscar}. 
\begin{itemize}
    \item \textbf{SAT}~\cite{xu2015show} uses grid-based deep visual features \cite{jiang2020defense} extracted by ResNet \cite{he2016deep}. Thus, we directly mask the objects, people, or both in the images using segmentation mask annotations, and feed the images into the captioning model to generate captions. 
    \item \textbf{OSCAR}~\cite{li2020oscar} leverages region-based deep visual features \cite{anderson2018bottom} extracted by a Faster-RCNN \cite{ren2016faster}. Therefore, instead of masking the objects, people, or both in the images, we remove the region-based features whose bounding box overlaps with the ground truth bounding by more than $50$ percent.
\end{itemize}

\section{List of gender-related words}
\label{sec:list-words}
We list the gender-related words that are replaced with the special token when inputting to gender classifiers:  
\textcolor{orange}{woman}, \textcolor{orange}{female}, \textcolor{orange}{lady}, \textcolor{orange}{mother}, \textcolor{orange}{girl}, \textcolor{orange}{aunt}, 
\textcolor{orange}{wife}, \textcolor{orange}{actress}, \textcolor{orange}{princess}, \textcolor{orange}{waitress}, \textcolor{orange}{sister}, \textcolor{orange}{queen}, \textcolor{orange}{pregnant}, \textcolor{orange}{daughter}, \textcolor{orange}{she}, \textcolor{orange}{her}, 
\textcolor{orange}{hers}, \textcolor{orange}{herself}, \textcolor{olive}{\textit{man}}, \textcolor{olive}{\textit{male}}, \textcolor{olive}{\textit{father}}, \textcolor{olive}{\textit{gentleman}}, \textcolor{olive}{\textit{boy}}, \textcolor{olive}{\textit{uncle}}, \textcolor{olive}{\textit{husband}}, \textcolor{olive}{\textit{actor}}, \textcolor{olive}{\textit{prince}}, \textcolor{olive}{\textit{waiter}}, \textcolor{olive}{\textit{son}}, \textcolor{olive}{\textit{brother}}, \textcolor{olive}{\textit{guy}}, \textcolor{olive}{\textit{emperor}}, \textcolor{olive}{\textit{dude}}, \textcolor{olive}{\textit{cowboy}}, \textcolor{olive}{\textit{he}}, \textcolor{olive}{\textit{his}}, \textcolor{olive}{\textit{him}}, \textcolor{olive}{\textit{himself}} and their plurals. \textcolor{orange}{Orange}/\textcolor{olive}{\textit{Olive}} denotes feminine/masculine words used to calculate Ratio, Error, BA, and $\text{DBA}_G$.

\section{Visual examples}
\label{sec:examples}

\begin{figure*}[t]
\centering
\vspace{10pt}
\includegraphics[clip, width=0.92\textwidth]{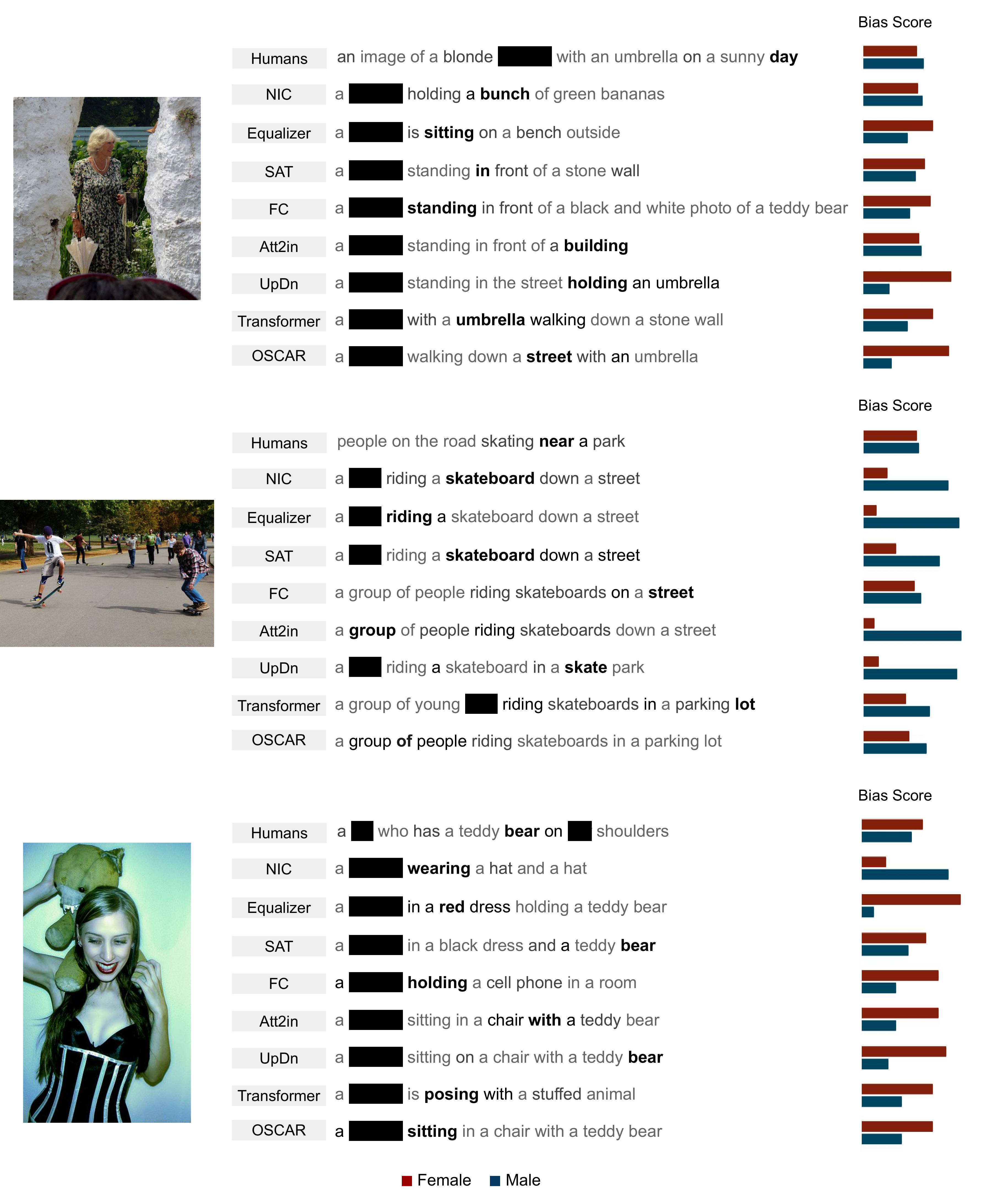}
% \vspace{-3pt} 
\vspace{10pt}
\caption{For each caption generated by humans or the models evaluated in the paper, we show our proposed bias score for \textit{female} and \textit{male} attributes. The contribution of each word to the bias score is shown in gray-scale (bold for the word with the highest contribution). Gender related words are masked during training and testing.}
\label{fig:all}
\vspace{10pt}
\end{figure*}

\begin{figure*}[t]
\centering
\vspace{10pt}
\includegraphics[width = 0.65\textwidth]{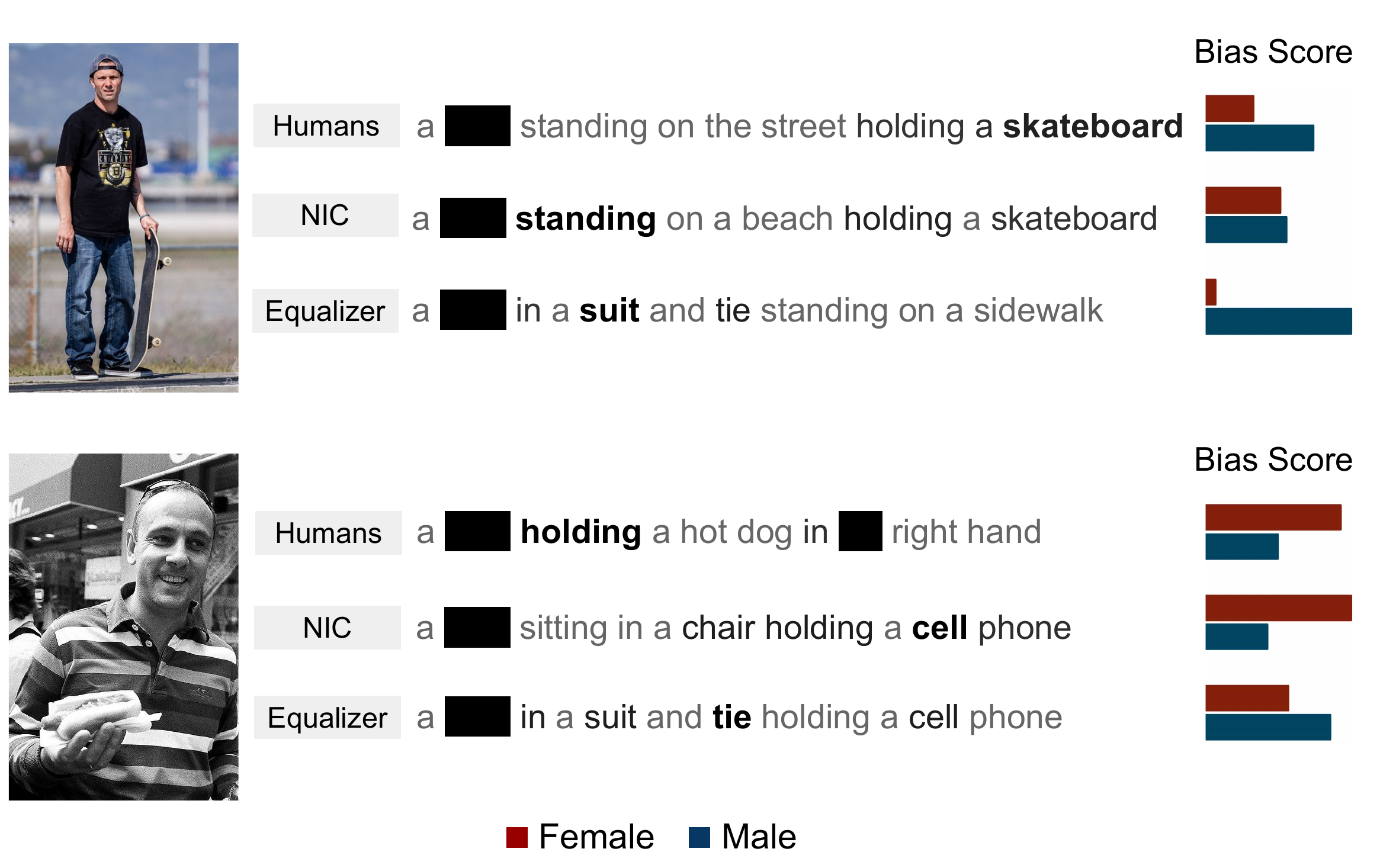}
\vspace{10pt}
\caption{Measuring gender bias in MSCOCO captions \cite{chen2015microsoft}. For each caption generated by humans, NIC \cite{vinyals2015show}, or NIC+Equalizer \cite{burns2018women}, we show our proposed bias score for \textit{female} and \textit{male} attributes. The contribution of each word to the bias score is shown in gray-scale (bold for the word with the highest contribution). Gender related words are masked during training and testing.}
\label{fig:malebias}
\end{figure*}

\begin{figure*}[t]
\centering
\vspace{10pt}
\includegraphics[width = 0.65\textwidth]{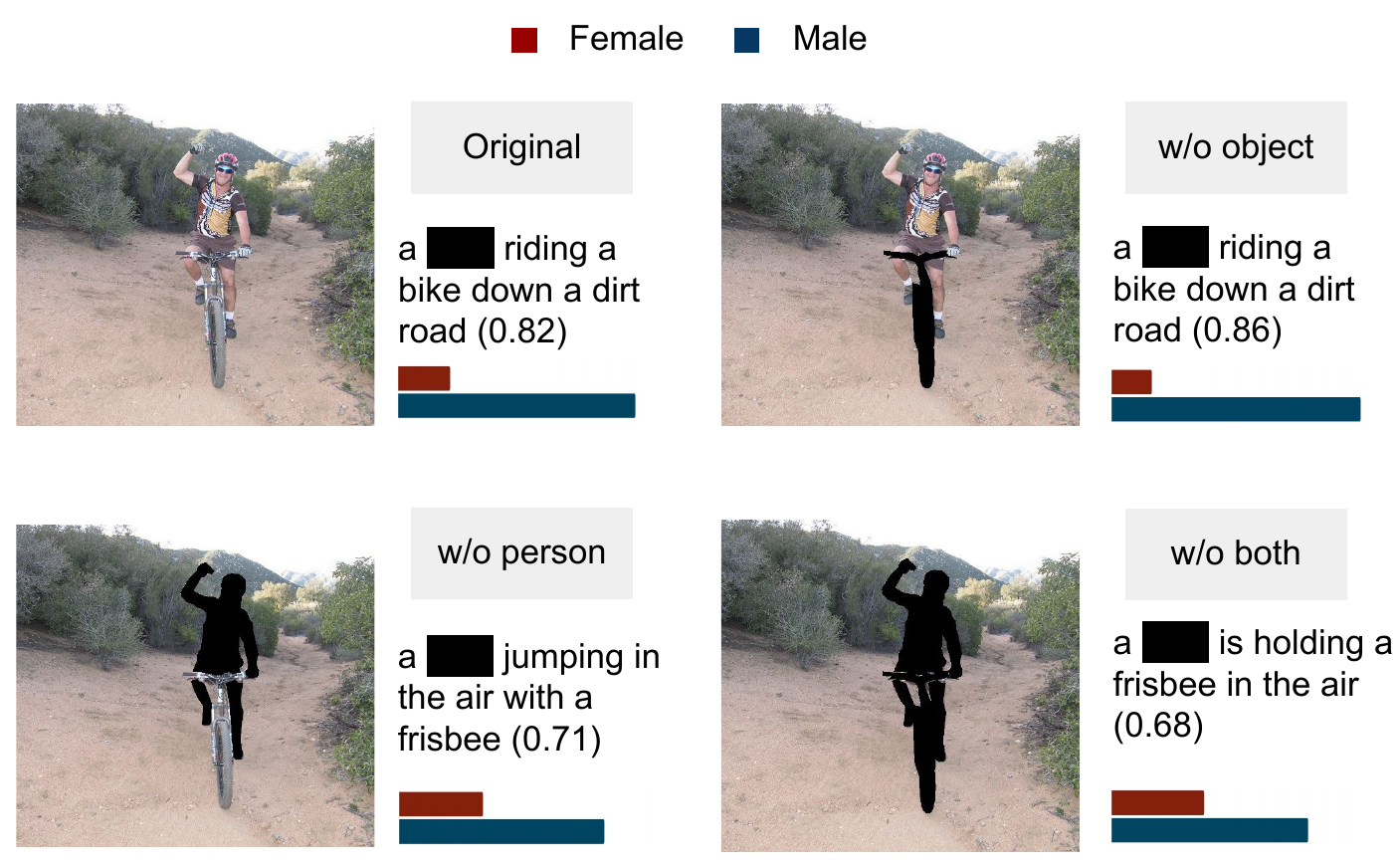}
\vspace{10pt} 
\caption{Generated captions and bias scores when images are partly masked.}
\label{fig:mask}
\end{figure*}

Here, we show more visual examples that could not be included in the main paper due to space limitations. Figure~\ref{fig:all} shows generated captions and their bias score for all the models evaluated in the main paper. Additionally, Figure~\ref{fig:malebias} shows more examples where NIC+Equalizer produces words strongly associated with gender stereotypes even when the evidence is not contained in the image. Whereas in the main paper we showed samples for women, here we show samples for men. It can be seen that NIC+Equalizer generates male-related words (\eg, \textit{suit}, \textit{tie}), and thus, obtain a higher bias score. We also show additional examples when images are partly masked in Figure~\ref{fig:mask}. The generated caption when the person (man) and the most correlated object (bicycle) are masked still contains a large bias score towards male.

%\section{Additional results}
%\label{sec:moreresults}

%In this section, we show the results of the additional experiments, which lead to deeper analysis.

%\subsection{BERT for race}
\section{Additional results}
\label{sec:moreresults}
We compare LIC for race when using different language encoders in Table~\ref{tab:race}. As with gender bias, the results show that LIC is consistent across different language models.

% \subsection{Difference between the protected attributes}
% \alert{Calculate the bias score for each protected attribute (gender, race)}

\section{Potential negative impact}
\label{sec:impact}

A potential negative impact of the use of the LIC metric to evaluate societal bias in image captioning is that researchers and computer vision practitioners may underestimate the bias and their impact in their models. Although it is important to have a tool to measure societal bias in computer vision models, we need to note that none metric can ensure the actual amount of bias. In other words, even if LIC (or any other metric) is small, or even zero, the model may still be biased. Therefore, relying on a single metric may overlook the problem. 

Additionally, whereas we use the value of LIC as the amount of bias amplification on a model, the definition of bias is different among existing work. As there is no standard definition of bias for image captioning, we should notice that our method is, perhaps, not the most appropriate one for all the contexts, and researchers should carefully consider which metric to use according to each application.

%%%%%%%%% REFERENCES
% \newpage

{\small
\bibliographystyle{ieee_fullname}
\bibliography{egbib}
}